\documentclass[11pt]{article}

\usepackage[preprint]{muset}

\usepackage{times}
\usepackage{latexsym}

\usepackage[T1]{fontenc}
\usepackage[utf8]{inputenc}

\DeclareUnicodeCharacter{202F}{\,}  % Narrow no-break space
\DeclareUnicodeCharacter{2076}{\textsuperscript{6}}  % Superscript 6

\usepackage{inconsolata}

\usepackage{graphicx}
\usepackage{amsmath}
\usepackage{tcolorbox}
\tcbuselibrary{breakable,skins}
\usepackage{listings}
\usepackage{enumitem}
\usepackage{hyperref}
\usepackage{tabularx}  % 放在导言区
\usepackage{booktabs}
\usepackage{array}

\usepackage{booktabs}
\usepackage{amssymb}
\usepackage{array}
\usepackage{xcolor}
\usepackage{longtable}
\usepackage{float}
\usepackage{caption}
\usepackage{adjustbox}
\usepackage[table]{xcolor}
\usepackage{booktabs}
\definecolor{drbHeader}{RGB}{240,240,245}
\definecolor{drbRow}{RGB}{248,248,252}
\definecolor{accent}{HTML}{8C6E54}

\definecolor{mygreen}{HTML}{7E97A6}
\definecolor{myred}{HTML}{C26B4A}
\colorlet{green}{mygreen}
\colorlet{red}{myred}
\definecolor{porcelain}{HTML}{F7F4EF}

\title{DeepResearch Bench II: Diagnosing Deep Research Agents via Rubrics from Expert Reports}

\author{%
\textbf{Ruizhe Li\textsuperscript{1,*}, Mingxuan Du\textsuperscript{1,*}, Benfeng Xu\textsuperscript{1,2,\ensuremath{\dagger}}, Chiwei Zhu\textsuperscript{1}, Xiaorui Wang\textsuperscript{2}, Zhendong Mao\textsuperscript{1,\S}}\\
\textsuperscript{1}University of Science and Technology of China\\
\textsuperscript{2}Metastone Technology, Beijing, China\\
\texttt{\{imlrz, dumingxuan\}@mail.ustc.edu.cn}\\
\textsuperscript{*}Equal contribution.\;
\textsuperscript{\ensuremath{\dagger}}Project lead.\;
\textsuperscript{\S}Corresponding author.%
}

\makeatletter
\@ifundefined{if@anonymous}{%
  \newif\if@anonymous
  \@anonymousfalse % 改成 \@anonymoustrue 则强制匿名
}{}
\makeatother

\makeatletter
\providecommand{\@toptitlebar}{}
\providecommand{\@bottomtitlebar}{}
\makeatother

\begin{document}
\maketitle

\begin{abstract}
Deep Research Agents (DRA) aim to help users search the web, synthesize information, and deliver comprehensive investigative reports. 
Prior benchmarks often either under-evaluate a system’s ability to produce meaningful insights and high-quality writing, or adopt coarse or LLM-defined criteria that are hard to verify and can diverge from human expert judgment.
To address these issues, we introduce \textbf{Deep Research Bench II}, a new benchmark for evaluating DRAs. It contains 132 grounded research tasks across \textbf{22} domains; for each task, an agent must produce a research report that is evaluated by a set of \textbf{9{,}430} fine-grained binary rubrics in total, covering three dimensions: information recall, analysis, and presentation. All rubrics are derived from carefully selected \textbf{expert-written investigative articles} and are constructed through a four-stage LLM+human pipeline that combines automatic extraction with over \textbf{400 human-hours} of expert review, ensuring that the criteria are verifiable and aligned with human expert judgment. 
We evaluate several state-of-the-art deep-research agents on Deep Research Bench II and find that even the strongest models satisfy fewer than 50\% of the rubrics, revealing a substantial gap between current DRAs and human experts. We release the benchmark, evaluation scripts, and all rubrics at \url{https://github.com/imlrz/DeepResearch-Bench-II} to facilitate future research on deep-research agents.
\end{abstract}

\begin{figure*}[h]
    \centering
    \includegraphics[width=\textwidth]{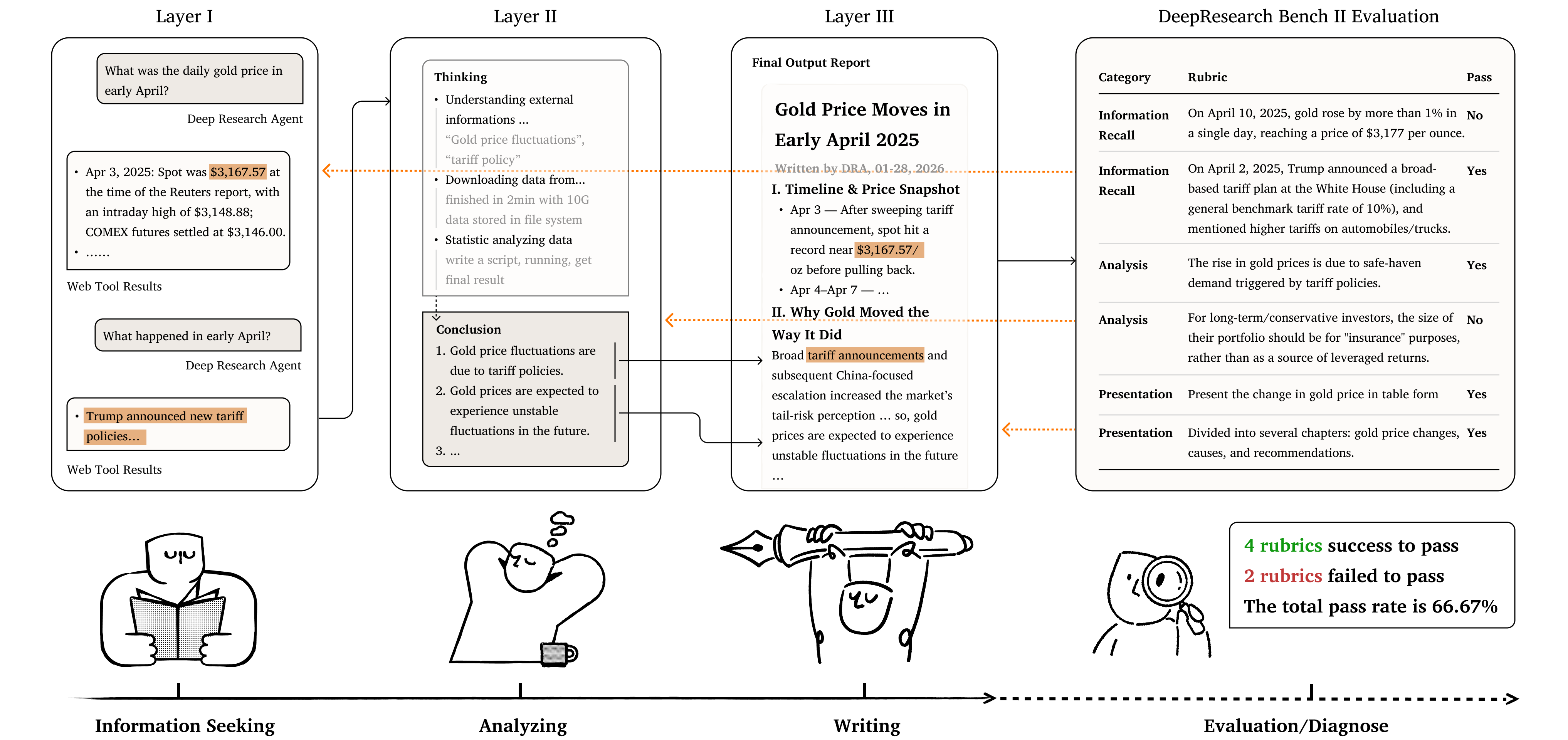}
    \caption{Illustration of the three-layer framework for Deep Research—information seeking, analyzing, and writing—and the corresponding three evaluation dimensions in our benchmark: information recall, analysis, and presentation.}
    \label{fig:intro}
\end{figure*}

\section{Introduction}

Deep Research Agents (DRA) are designed to help users tackle complex, open-ended information needs by searching the web, synthesizing heterogeneous evidence, and delivering comprehensive investigative reports. In practice, most deployed DRAs are instantiated as LLM-based agents that orchestrate language understanding, planning, and tool use (e.g., web search and browsing) to complete research-style tasks \citep{brown2020languagemodelsfewshotlearners,andreas2022languagemodelsagentmodels,nakano2022webgptbrowserassistedquestionansweringhuman}. Recent commercial systems such as Gemini Deep Research and OpenAI Deep Research explicitly target multi-step online investigation: they decompose user queries, explore large numbers of web sources, and produce analyst-level reports that aggregate and interpret retrieved information \citep{gemini_deep_research,openai_deep_research}. Despite these advances, current DRAs still show clear limitations in both information seeking and reasoning: they frequently miss key sources, over-rely on a small subset of retrieved documents, and struggle to form stable, well-justified viewpoints from conflicting evidence \citep{White2024Advancing}. This gap motivates rigorous, fine-grained evaluation frameworks that can reveal where DRAs truly fall short relative to human experts.

\begin{figure*}[h]
    \centering
    \includegraphics[width=0.85\textwidth]{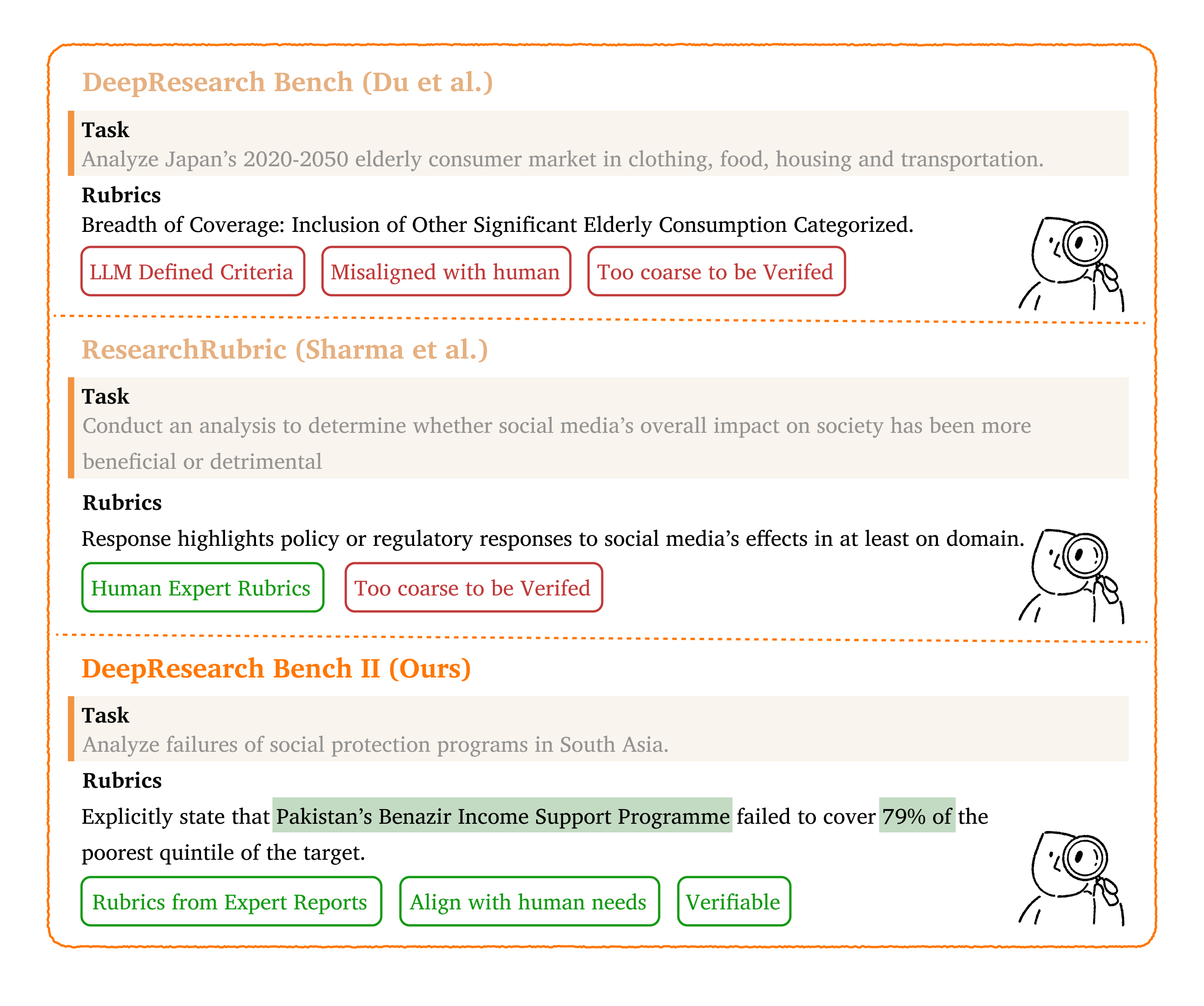}
    \caption{
Comparison of evaluation schemes in prior deep-research benchmarks and \textbf{DeepResearch Bench II}. 
\emph{Top}: Benchmarks that rely on LLM-defined criteria can be misaligned with human experts. 
\emph{Middle}: Benchmarks that adopt human-written but coarse rubrics allow seemingly correct hallucinations from the DRA to pass. 
\emph{Bottom}: DeepResearch Bench II derives fine-grained, content-bearing rubrics from human expert reports, enabling the LLM judge to reject seemingly correct hallucinations and provide unbiased, verifiable evaluations aligned with human judgment.
}
    \label{fig:compare}
\end{figure*}

To systematically assess DRA capabilities, several deep-research benchmarks have been proposed. Broadly, they fall into two categories. The first focuses on fixed-answer tasks, where the agent must retrieve specific entities or numerical facts from the web \citep{wei2025browsecompsimplechallengingbenchmark,chen2025xbenchtrackingagentsproductivity,futuresearch2025deepresearchbenchevaluating,wong2025widesearchbenchmarkingagenticbroad}. These setups capture whether an agent can locate and extract relevant information, but they only partially reflect real-world user needs: they rarely test whether the agent can decide \emph{what} to look for under open-ended goals, or how to integrate findings into a coherent narrative. The second category evaluates full research reports, typically along dimensions such as comprehensiveness, insight, and citation quality \citep{du2025deepresearchbenchcomprehensivebenchmark,xu2025researcherbenchevaluatingdeepai,fan2025understandingdeepresearchreports}. However, these benchmarks suffer from structural issues: their evaluation criteria are often defined directly by LLMs, which can introduce systematic misalignment with human expert judgments; moreover, the rubrics are overly coarse and weakly interpretable, and may require judge LLMs to rely on out-of-distribution or unverifiable internal knowledge, leading to inaccurate and hard-to-trust scores (as illustrated in Figure~\ref{fig:compare}).

To address these limitations, we introduce \textbf{Deep Research Bench II}, a new benchmark for evaluating Deep Research Agents (DRA). Starting from the domain distribution of DeepResearch Bench, we collect high-quality, expert-written investigative reports from reputable open-access venues. We then construct research-style tasks that mirror these domains and filter them through a series of quality checks, resulting in a dataset of 132 tasks across 22 domains.

On top of these source articles, we build a four-stage pipeline to extract and refine evaluation criteria. Using LLM-based extraction, self-evaluation filtering, manual cleaning, and domain-expert refinement, we obtain 9{,}430 fine-grained binary rubrics in total. Each rubric is derived directly from expert reports and encodes a concrete factual or inferential requirement. We further organize all rubrics into three dimensions that correspond to the core capabilities of DRAs: \emph{Information Recall} (whether the agent retrieves the right evidence), \emph{Analysis} (whether it produces meaningful higher-level insights), and \emph{Presentation} (whether it structures and communicates the report in a clear way). Given a model-generated report, an LLM judge evaluates in an end-to-end manner whether each rubric is satisfied, providing dimension-wise scores for every task.

We use Deep Research Bench II to benchmark a diverse set of state-of-the-art deep research agents, including both open-source and closed-source frontier models. The results reveal a clear and consistent gap between current DRAs and human experts: even the strongest agents fail to pass more than 50\% of the rubrics, with especially large deficits in Information Recall and Analysis. We further conduct human–LLM agreement studies to validate the robustness of our evaluation protocol and analyze where existing models systematically fall short, with the goal of guiding future progress in deep research.

In summary, our contributions are threefold:
\begin{itemize}
    \item We introduce a grounded, expert-derived benchmark for deep research with 132 tasks and 9{,}430 verifiable rubrics constructed from real expert reports.
    \item We propose a three-dimensional evaluation framework and an LLM-as-judge protocol that jointly assess information recall, analysis, and presentation in a fine-grained, rubric-based manner.
    \item We conduct a comprehensive empirical study of leading DRAs, quantify their gap to human experts, and provide analysis and human-alignment experiments that set a new reference point for future deep-research evaluation.
\end{itemize}

\section{Related Work}

\subsection{Deep Research Benchmarks}

BrowseComp \citep{wei2025browsecompsimplechallengingbenchmark}, XBench \citep{chen2025xbenchtrackingagentsproductivity}, Deep Research Bench \citep{futuresearch2025deepresearchbenchevaluating}, WideSearch \citep{ wong2025widesearchbenchmarkingagenticbroad} focus on tasks that have fixed answers, such as specific entities or numbers. These benchmarks primarily assess whether the agent can locate and retrieve the correct information, testing the agent’s abilities in planning, reasoning, and discerning information sources. 

On the other hand, tasks in DeepResearch Bench\citep{du2025deepresearchbenchcomprehensivebenchmark}, Researcher Bench \citep{xu2025researcherbenchevaluatingdeepai}, DeepResearch-ReportEval \citep{fan2025understandingdeepresearchreports}, LiveResearchBench \citep{wangLiveResearchBenchLiveBenchmark2025}, and ResearchRubrics \citep{sharmaResearchRubricsBenchmarkPrompts2025} simulate user queries and require the Deep Research agent to deliver a complete research report. These benchmarks evaluate the quality of the final report based on specific dimensions such as comprehensiveness, insight, and citation accuracy. Compared to the first category, these open-ended benchmarks are better at assessing whether the Deep Research agent knows which information should be recalled and its ability to analyze the information. However, they also have their own issues. In DeepResearch Bench, DeepResearch-ReportEval, and LiveResearchBench, the specific evaluation criteria are defined by the LLM itself, with human expert involvement only during the review phase or final consistency assessment. This approach may introduce anchoring bias \citep{du2025deepresearchbenchcomprehensivebenchmark, fan2025understandingdeepresearchreports, wangLiveResearchBenchLiveBenchmark2025, sharmaResearchRubricsBenchmarkPrompts2025}. In contrast, the evaluation criteria designing in ResearcherBench and ResearchRubrics involve human expert, which can better reflect human expert cognition to some extent \citep{xu2025researcherbenchevaluatingdeepai, sharmaResearchRubricsBenchmarkPrompts2025}. However, the granularity of their rubrics is insufficient, and correctly evaluating them usually requires the model having specific internal knowledge. It is also noteworthy that DeepResearch Bench, DeepResearch-ReportEval, LiveResearchBench, and Researcher Bench evaluate whether the citations support the claims, but this only measures the accuracy of the citations and does not necessarily confirm that the information is correct (e.g., false information from unofficial sources) \citep{du2025deepresearchbenchcomprehensivebenchmark, fan2025understandingdeepresearchreports, wangLiveResearchBenchLiveBenchmark2025, xu2025researcherbenchevaluatingdeepai}. Essentially, these benchmarks still do not verify the accuracy of recalled information in the same way as the first category benchmarks. 

\begin{table*}[t]
    \centering
    \small
    \begin{tabular}{l*{4}{>{\centering\arraybackslash}p{2.5cm}}}
    \toprule
    \textbf{Method} & \textbf{Real-world Topics} & \textbf{Rubrics source from Human Experts} & \textbf{Verifiable by LLM Internal Knowledge} & \textbf{Average \# Rubrics per task} \\
    \midrule
    BrowseComp et al. & \textcolor{red}{$\boldsymbol{\times}$} & \textcolor{green}{$\boldsymbol{\checkmark}$} & \textcolor{green}{$\boldsymbol{\checkmark}$} & - \\
    DeepResearch Bench et al. & \textcolor{green}{$\boldsymbol{\checkmark}$} & \textcolor{red}{$\boldsymbol{\times}$} & \textcolor{red}{$\boldsymbol{\times}$} & 10-25 \\
    Researcher Bench & \textcolor{red}{$\boldsymbol{\times}$} & \textcolor{green}{$\boldsymbol{\checkmark}$} & \textcolor{red}{$\boldsymbol{\times}$} & 14 \\
    ResearchRubrics & \textcolor{green}{$\boldsymbol{\checkmark}$} & \textcolor{green}{$\boldsymbol{\checkmark}$} & \textcolor{red}{$\boldsymbol{\times}$} & 26 \\
    \midrule
    OURS & \textcolor{green}{$\boldsymbol{\checkmark}$} & \textcolor{green}{$\boldsymbol{\checkmark}$} & \textcolor{green}{$\boldsymbol{\checkmark}$} & 71 \\
    \bottomrule
    \end{tabular}
    \caption{Comparison of different methods based on key criteria: whether they use real-world topics, source rubrics from human experts, allow verification by LLM’s internal knowledge, and Average \# Rubrics per task. 'OURS' stands out for meeting all criteria, emphasizing its comprehensive approach. BrowseComp et al. includes BrowseComp, XBench, Deep Research Bench, and WideSearch. DeepResearch Bench et al. includes DeepResearch Bench, DeepResearch-ReportEval, and LiveResearch-Bench.}
    \label{tab:comparison_methods}
    \end{table*}

\subsection{LLM as judge and Rubric-based Evaluation}

With its strong language comprehension abilities, LLMs are well-suited for handling evaluations in natural language, and LLM as judge has become a common evaluation method in NLP and various other fields \citep{bavaresco2025llmsinsteadhumanjudges,gu2025surveyllmasajudge,zheng2023judgingllmasajudgemtbenchchatbot}. However, research has shown that LLMs, as judges, still struggle to perform well on certain complex and open-ended tasks \citep{li2025automatedcreativityevaluationlarge,chakrabarty2024artartificelargelanguage,marioriyad2025silentjudgeunacknowledgedshortcut,thakur2025judgingjudgesevaluatingalignment}. Recently, rubrics have been widely adopted for these types of evaluation tasks \citep{aroraHealthBenchEvaluatingLarge2025,lin2024wildbenchbenchmarkingllmschallenging,sirdeshmukh2025multichallengerealisticmultiturnconversation,starace2025paperbenchevaluatingaisability, gou2025mind2web2evaluatingagentic}. By breaking down tasks into smaller scoring points, rubrics make the evaluation process by LLM as judge more interpretable. Rubrics designed by human experts also help ensure the consistency of LLM judge scores with human expert preferences to some extent \citep{pathak2025rubricneedenhancingllmbased, Hashemi_2024}. In this context, the rubric-based evaluation method has proven effective. Compared to direct scoring, rubrics introduce external knowledge to LLMs - namely, human understanding of "how this task should be broken down."

Although the term "rubric" is used in both cases, rubrics designed by different works vary in granularity, which places different demands on the evaluation model. For instance, in HealthBench, there is a rubric like: "Briefly describes common causes of muscle weakness in infants." \citep{aroraHealthBenchEvaluatingLarge2025} Although it breaks a complex medical task into such a small point, verifying this rubric requires the judge model to have relevant medical knowledge. For models without this knowledge, even if the evaluation subject provides incorrect answers, they will not be detected. Having relevant domain knowledge is feasible in fields such as healthcare and programming, and current research is working to enable LLMs to achieve this \citep{huynh2025largelanguagemodelscode, Singhal2023LargeLM}. However, for deep research tasks — which inherently cannot be verified by the model’s internal knowledge — if the rubric is still like the one mentioned above, the rubric can only validate the agent's planning ability and report organization (since the rubric decomposes the answer content and proves that the response includes the point). It cannot validate the accuracy of the information, and thus cannot assess the search and analysis capabilities of the deep research agent.

\section{Methodology}

\subsection{Task Collection and Rubric Design}

\subsubsection{Source Article Selection}

\begin{figure*}[h]
    \centering
    \includegraphics[width=0.8\textwidth]{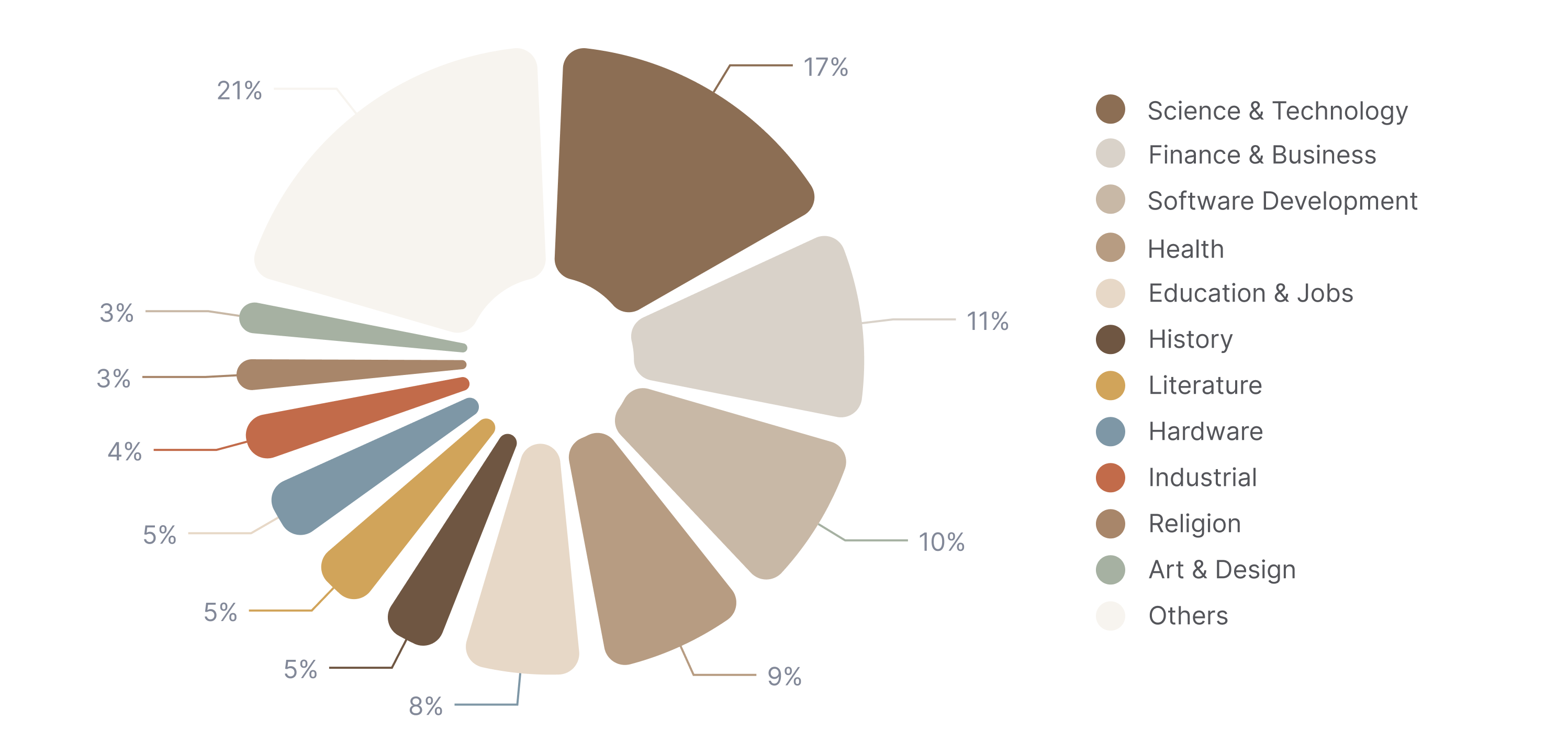}
    \caption{Topic distribution of source articles/tasks used for our benchmark.}
    \label{fig:topic-distribution}
\end{figure*}

DeepResearch Bench analyzed the topic distribution of real-world user queries and invited domain experts to design benchmark tasks aligned with these distributions \citep{du2025deepresearchbenchcomprehensivebenchmark}. Building upon this foundation, we used the tasks from DeepResearch Bench as seeds and searched for open-access review reports addressing similar or identical research questions in reputable journals, top conferences, and credible institutional publications (commercial or governmental). These reports are typically written by domain experts over the course of weeks or months and undergo multiple layers of validation---from peer reviewers, editors, practitioners, and the general public---thus reflecting a high degree of credibility and expert consensus.

For each collected report, we verified its license and retained only those under CC-4.0 or CC-4.0-NC terms, which permit non-commercial use and adaptation. From this pool, we manually selected articles suitable for deep research tasks based on four criteria:
(1) the work requires extensive open-source information gathering;
(2) the conclusions involve substantial synthesis and analysis;
(3) the findings do not depend on physical experiments (e.g., in chemistry or biology); and
(4) the analyses are reproducible without reliance on complex data-mining pipelines or stochastic machine learning models.
After filtering, 132 articles were retained (see Appendix~\ref{sec:source-article-list} for the full list). These serve as the source materials for constructing our tasks and extracting the ground-truth rubrics for evaluation.

\subsubsection{Task and Rubric Design Protocol}
\label{sec:task-rubric-requirements}

\begin{table}[h]
  \centering
  \small
  % 交替行底色：从第二行开始用 accent!5
  \rowcolors{2}{accent!5}{white}
  \begin{tabular}{p{0.18\textwidth}p{0.75\textwidth}}
    % 表头行：底色浅棕，文字用主色
    \rowcolor{accent!15}
    \textbf{\textcolor{accent}{Dimension}} & \textbf{\textcolor{accent}{Description}} \\
    \toprule
    \textbf{Information Recall} & Evaluates whether the model can use planning and reasoning, together with search tools, to accurately and comprehensively retrieve task-relevant information from the open web. This includes identifying what information needs to be collected, locating it from large and heterogeneous sources, and checking its reliability through source credibility and cross-validation. \\
    \addlinespace[0.4em]
    \textbf{Analysis} & Evaluates whether the model can synthesize the retrieved information and derive higher-level insights that go beyond simple aggregation. Typical expectations include extracting trends or paradigms, forming causal or comparative explanations, and drawing conclusions that directly address the research question rather than merely restating evidence. \\
    \addlinespace[0.4em]
    \textbf{Presentation} & Evaluates whether the model can organize and communicate its findings in a clear, user-accessible, and verifiable way. This covers coherent structure, appropriate use of formats such as tables, figures, and lists, transparent citation or referencing for verification, and a level of explanation that is compatible with the plausible background of the target user. \\
    \bottomrule
  \end{tabular}
  \caption{Three evaluation dimensions used in DeepResearch Bench II.}
  \label{tab:dimensions}
\end{table}

For each task derived from a source article, we impose two basic requirements. 
First, the task may either correspond to the core research question of the source article or cover a relevant subset, but it must \emph{always} require both information collection and non-trivial analysis, rather than pure fact lookup. 
Second, for time-sensitive topics (e.g., scientific progress, industry dynamics), the task must explicitly specify the temporal scope of the investigation to ensure consistency with the time period covered by the source article.

For evaluation, we adopt a full-rubric scoring scheme. 
Each rubric is a binary criterion, typically phrased as a concrete factual or inferential requirement (e.g., “Presented the change in gold prices in table form”). 
A rubric is marked as passed only if the report satisfies the specified requirement; the final task score is computed as the proportion of passed rubrics. 
Dimension-wise scores are obtained by aggregating rubrics assigned to the corresponding dimension in Table~\ref{tab:dimensions}.

To ensure that rubrics are both verifiable and fine-grained across all three dimensions, we impose the following constraints:
(1) each rubric must capture information or reasoning that is essential for correctly answering the task;
(2) each rubric must be atomic and indivisible, i.e., complex statements should be split into smaller rubrics, each checking a single fact or inference;
(3) each rubric should directly encode the target content rather than only specifying a broad topic (e.g., “stated that labor loss in small cities is due to job-structure mismatch” rather than “gave reasons for labor loss”);
and (4) for numerical data, each value must be explicitly verified through a rubric, with an acceptable margin of error allowed only for quantities that require intermediate computation.

To capture the different facets of deep research performance in a structured way, we decompose each task into three evaluation dimensions and assign rubrics accorddingly. 
Table~\ref{tab:dimensions} summarizes these dimensions.

\subsubsection{Task and Rubric Construction Pipeline}

\begin{figure*}[t]
    \centering
    \includegraphics[width=0.8\textwidth]{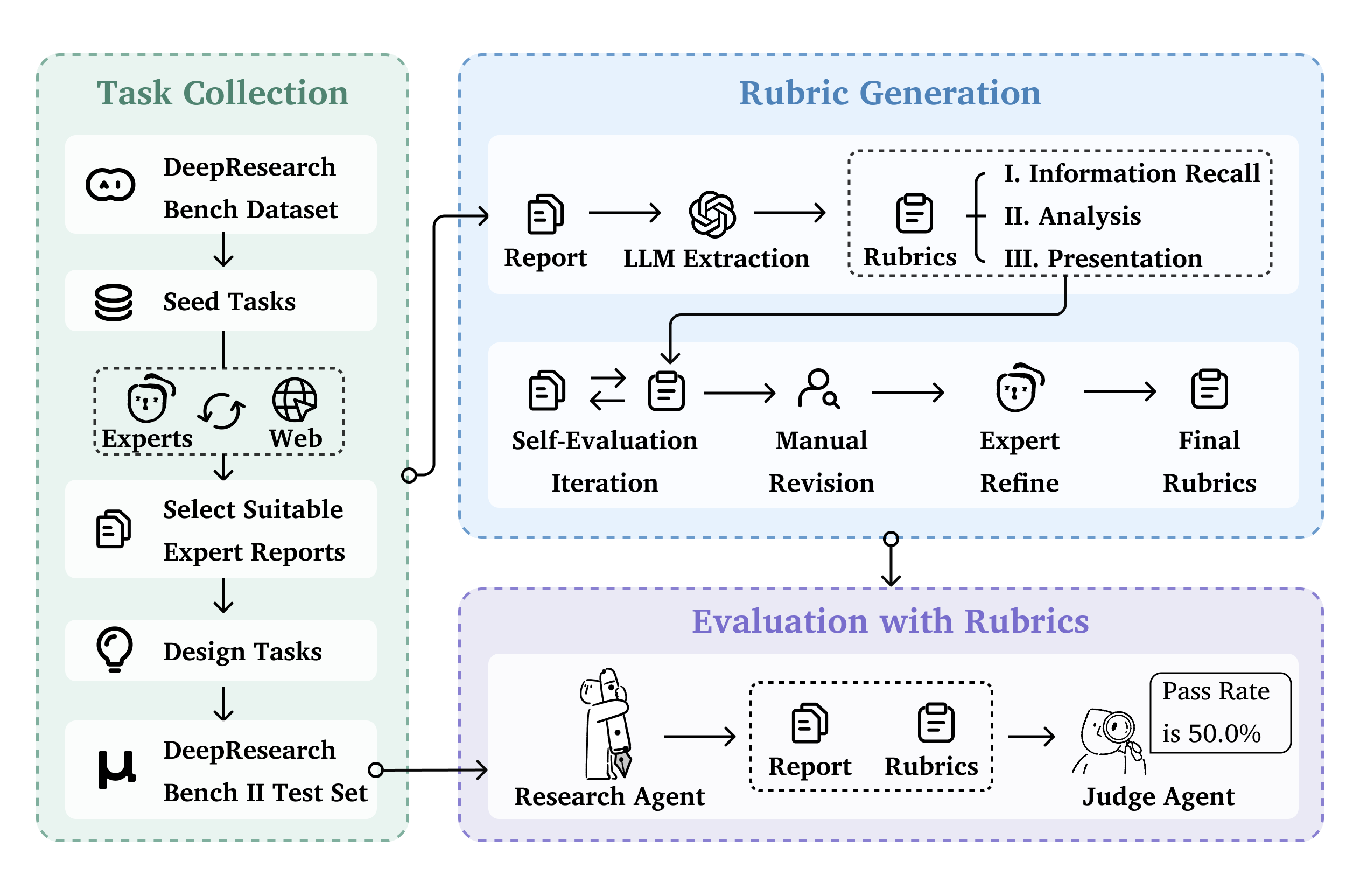}
    \caption{This diagram illustrates the entire workflow of our work. By decomposing human-expert articles into fine-grained, verifiable rubrics, we enable a comparison between human-expert articles and model-generated articles.}
    \label{fig:method}
    \end{figure*}

The construction of tasks and rubrics proceeds in four stages:  
(1) \textbf{LLM extraction}, (2) \textbf{self-evaluation iteration}, (3) \textbf{manual revision}, and (4) \textbf{expert review and refinement}.

\paragraph{LLM Extraction}
We directly employ a large language model (denoted as \textit{task-generator}) to extract tasks and corresponding rubrics from each source article. Carefully designed prompts guide the model to ensure that the outputs meet the requirements defined in Section~\ref{sec:task-rubric-requirements}---covering both information recall, analysis and presentation, producing verifiable, fine-grained rubrics.

\paragraph{Self-Evaluation Iteration}
Since LLM generations are prone to hallucinations \citep{simhi2025trustmeimwrong, Huang_2025}---especially in long-context settings \citep{ Farquhar2024DetectingHL, liu2025longcontexthallucinationdetection}---we adopt a \textit{self-evaluation} strategy to mitigate this issue. Each initially generated rubric is applied to evaluate its own source article using the evaluation script.  
If the resulting \textit{Information Recall} or \textit{Analysis} dimension achieves an accuracy below 90\%, the generation is considered unreliable due to potential hallucination, and the LLM is prompted again to regenerate the task and rubrics until the threshold is met.

\paragraph{Manual Revision}
After the self-evaluation stage, human annotators manually inspect all generated tasks and rubrics to ensure compliance with the requirements in Section~\ref{sec:task-rubric-requirements}. This step primarily eliminates logically inconsistent or redundant rubrics and refines ambiguous phrasing to improve clarity and precision.

\paragraph{Expert Review and Refinement}
Because LLMs may lack domain-specific expertise, the automatically generated tasks and rubrics can still contain residual issues after self-evaluation, such as including rubrics that, while consistent with the source article, are not essential for answering the defined task, retaining subtle hallucinations, or omitting critical criteria needed for a faithful evaluation of the task. To mitigate these problems, we recruited \textbf{domain experts} across the fields represented in our benchmark. Over the course of more than \textbf{400 human-hours}, these experts conducted in-depth review, discussion, and iterative refinement of all tasks and rubrics. During this process, they ensured that rubrics in the \emph{Information Recall} dimension correspond to essential or genuinely supportive information, remain fully consistent with the source article, and rely only on publicly accessible data. For the \emph{Analysis} dimension, they required that each rubric encode reasoning or synthesis rather than direct factual lookup, that every analytical conclusion meaningfully addresses the task question, and that conclusions are logically grounded in the evidence presented. For the \emph{Presentation} dimension, they checked that rubrics reflect any explicit formatting or structural requirements stated in the task and that they collectively encourage reports that are coherent, persuasive, and accessible to the intended user. This expert curation step is crucial for aligning the rubric set with human evaluation standards and for improving both the validity and reliability of our benchmark.

\noindent
This four-stage pipeline ensures that each task and rubric is not only grounded in expert-authored literature but also verifiable, fine-grained, and aligned with human evaluation standards.
  
\section{Experiment}
\label{sec:experiment}

\subsection{Experimental Setup}

\subsubsection{Implementation Details}
For all evaluation tasks, we employ \texttt{Gemini-2.5-Pro} as the Judge LLM. 
This model is selected due to its strong reasoning capability and stable binary classification performance. 
All scoring tasks are conducted in batches of 50 rubrics per evaluation pass, a configuration chosen to balance accuracy and computational efficiency. 
Each rubric is scored independently using a binary scheme (0/1), and the final score for a task is computed as the fraction of rubrics marked as passed.

\subsubsection{Evaluated Models}
We benchmark a diverse suite of Deep Research agents and leading LLM-based agents. 
Specifically, our evaluated models include: OpenAI-GPT-o3 Deep Research \citep{OpenAI_DeepResearch_2025}, Gemini-3-Pro Deep Research \citep{haas2025build-gemini-deep-research}, Gemini-2.5-Pro Deep Research \citep{Google_Gemini_DeepResearch_2024}, Grok Deep Search, Perplexity Research \citep{Perplexity_DeepResearch_2025}, Qwen3-Max Deep Research \citep{Alibaba_Qwen_DeepResearch_2025}, Doubao Deep Research and Tongyi Deep Research \citep{tongyidr}. 
These agents represent the latest commercially deployed or early-released Deep Research agents available as of November~2025. 

\subsection{Main Results}

\begin{table*}[t]
  \centering
  \begin{tabularx}{\textwidth}{l *{4}{>{\centering\arraybackslash}X}}
  \toprule
  \textbf{Model} & \textbf{InfoRecall} & \textbf{Analysis} & \textbf{Presentation} & \textbf{TotalScore} \\
  \midrule
  OpenAI-GPT-o3 Deep Research \citep{OpenAI_DeepResearch_2025}        & \textbf{39.98} & \underline{49.85} & 89.16 & \textbf{45.40} \\
  Gemini-3-Pro Deep Research \citep{haas2025build-gemini-deep-research}        & \underline{39.09} & 48.94 & \textbf{91.85} & \underline{44.60} \\
  Gemini-2.5-Pro Deep Research \citep{Google_Gemini_DeepResearch_2024}        & 34.91 & \textbf{51.91} & 90.24 & 41.98 \\
  Doubao Deep Research \citep{bytedance_doubao_chat}       & 34.83 & 49.43 & 83.51 & 40.99 \\
  Qwen3-Max Deep Research \citep{Alibaba_Qwen_DeepResearch_2025}         & 34.18 & 48.04 & 74.59 & 39.25 \\
  Grok Deep Search \citep{xai_grok}       & 33.52 & 42.50 & \underline{91.42} & 39.23 \\
  Perplexity Research \citep{Perplexity_DeepResearch_2025}  & 33.05 & 44.47 & 79.34 & 38.58 \\
  Tongyi Deep Research \citep{tongyidr}       & 22.95 & 35.89 & 86.13 & 29.89 \\
  \bottomrule
  \end{tabularx}
  \caption{Main evaluation results across all tested models on the DeepResearch Bench II. \textbf{Bold} and \underline{underlined} values indicate the best and second-best performance in each dimension, respectively.}
  \label{tab:mainresults}
  \end{table*}

Table~\ref{tab:mainresults} reports the performance of all evaluated Deep Research Agents across the three rubric dimensions---\textit{Analysis}, \textit{InfoRecall}, and \textit{Presentation}---as well as their total score. The results reveal substantial disparities in the agents' ability to execute long-context synthesis, structured reasoning, and task-aligned report generation.

\paragraph{Overall Performance.}
OpenAI-GPT-o3 Deep Research emerges as the strongest overall agent, achieving the highest scores in both Information Recall and the overall aggregated metric. Gemini-3-Pro and Gemini-2.5-Pro Deep Research stands out in the Information Recall and Analysis dimension and also performs competitively in Presentation, making it a solid runner-up with a more analysis-oriented profile. Grok Deep Search excels in Presentation, clearly leading in how it structures and communicates information, but its retrieval and reasoning capabilities lag behind the top agents. Tongyi Deep Research consistently ranks at the bottom across dimensions, reflecting a noticeable gap from the other evaluated models. Overall, OpenAI-GPT-o3 appears the most well-rounded, Gemini-3-Pro shows balanced strengths, and Grok Deep Search is highly polished in presentation but weaker in the upstream stages of deep research. However, even the best-performing agent still fails to satisfy more than half of the rubrics, indicating a substantial remaining gap between current DRAs and human experts.

\paragraph{Dimension-Level Insights.}
Across \textbf{InfoRecall}, OpenAI-GPT-o3 Deep Research and Gemini-3-Pro Deep Research clearly outperform the rest, while Gemini-2.5-Pro, Doubao Deep Research, Qwen3-Max Deep Research, Grok Deep Search, and Perplexity Research form a middle cluster with comparable retrieval performance. Tongyi Deep Research is noticeably weaker in this dimension, suggesting limitations in its information-seeking behavior. In \textbf{Analysis}, OpenAI-GPT-o3 and Gemini-2.5-Pro form the top tier, indicating stronger abilities to synthesize evidence and generate higher-level conclusions; Perplexity Research, Grok Deep Search and Gemini-3-Pro follow behind, with Tongyi Deep Research again trailing. For \textbf{Presentation}, Grok Deep Search and Gemini-3-Pro take a slight lead, with Gemini-2.5-Pro and OpenAI-GPT-o3 closely behind; all three produce generally well-structured and readable reports. Tongyi Deep Research and Perplexity Research perform somewhat worse in this dimension, pointing to less effective organization and user-facing communication of their findings.

\subsection{Ablation Study}
\label{sec:ablation}
To validate the design choices in our evaluation pipeline, we conduct controlled ablations along two critical axes: rubric batch size and evaluator model selection. 
We curate a set of ten model-generated reports covering diverse tasks, model families, and output formats (PDF, DOCX, Markdown, and HTML). 
Human annotators label these reports against the corresponding rubrics, and we measure the alignment between LLM-as-judge predictions and human judgments. 
All reported results reflect accuracy (ACC) and F1 scores computed against these human annotations.

\subsubsection{Rubric Batch Size}

\begin{table*}[h]
  \centering
  \begin{tabular}{lccc}
    \toprule
    \textbf{batch size} & \textbf{cost} & \textbf{ACC} & \textbf{F1} \\
    \midrule
    full & 0.2025 & 90.80 & 87.06 \\
    80   & 0.2087 & 91.47 & 88.36 \\
    50   & 0.2513 & 91.75 & 89.57 \\
    10   & 0.4832 & 92.29 & 89.37 \\
    5    & 0.8298 & 90.39 & 85.66 \\
    \bottomrule
    \end{tabular}
    \captionof{table}{Ablation on rubric batch size. The table shows the average evaluation cost per task (in dollars), accuracy (ACC), and F1-score for different batch sizes. Batch size 50 achieves the best F1-score while maintaining reasonable cost, making it the optimal choice for our evaluation pipeline.}
    \label{tab:batchsize}
  \end{table*}

We vary the number of rubrics evaluated per pass and compare accuracy, F1, and cost efficiency. 
Gemini-2.5-Pro is used as the evaluator for all configurations. 
The results are summarized in Table~\ref{tab:batchsize}.

We observe a consistent cost reduction as batch size increases. 
Accuracy and F1 initially improve and then plateau or slightly drop when the batch becomes too small, consistent with observations in prior studies \citep{wang2025evaluatingllmsmultipleproblems}. 
Table~\ref{tab:batchsize} illustrates that a batch size of 50 offers the best balance, delivering strong accuracy while keeping evaluation overhead manageable. 
Therefore, we adopt 50 as the default configuration.

\subsubsection{Evaluator Model Selection}
We further compare different LLMs as evaluators under a fixed batch size of 50. 
Each model evaluates the same set of human-labeled reports, and the agreement with human judgments is shown in Table~\ref{tab:evaluator}.

\begin{table}[h]
    \centering
    \begin{tabular}{lcc}
    \toprule
    \textbf{model} & \textbf{ACC} & \textbf{F1} \\
    \midrule
    Gemini-2.5-Pro   & 91.75 & 89.57 \\
    GPT-5            & 88.36 & 78.28 \\
    Gemini-2.5-Flash & 89.45 & 79.79 \\
    \bottomrule
    \end{tabular}
    \captionof{table}{Ablation on evaluator model choice.}
    \label{tab:evaluator}
\end{table}

Gemini-2.5-Pro demonstrates the highest alignment with human annotations across both metrics. 
GPT-5 and Gemini-2.5-Flash show noticeably lower consistency, particularly in F1. 
Given its superior reliability as an automatic judge, we select Gemini-2.5-Pro as the evaluator throughout all experiments.

\section{Analysis and Discussion}

\subsection{Robustness Analysis Across Language and Topic}

To evaluate model robustness, we analyze performance variations across languages (English vs. Chinese) and research topics (e.g., Health, Finance \& Business, Software Development). We employ \textit{relative scores}---defined as the deviation of each model's score from the task-average score---to control for inherent task difficulty differences. One-way ANOVA tests are conducted separately for language and topic factors.

\paragraph{Language Effects.}
Most models exhibit robust performance across languages, with p-values $>$ 0.05 for \textit{OpenAI}, \textit{Gemini}, \textit{Doubao}, \textit{Qwen}, and \textit{Perplexity}. Only \textit{Grok} shows a marginally significant difference (p = 0.034), suggesting a slight sensitivity to language variation.

\begin{table*}[h]
  \centering
  \small
  \begin{tabular}{lcccccccc}
  \toprule
  \textbf{Language} & \textbf{OpenAI} & \textbf{Gemini} & \textbf{Doubao} & \textbf{Qwen3} & \textbf{Grok} & \textbf{Perplexity} & \textbf{Tongyi} & \textbf{Avg.} \\
  \midrule
  en & 45.53 & 41.83 & 40.33 & 36.95 & 36.79 & 38.03 & 29.22 & 38.38 \\
  zh & 45.26 & 42.13 & 41.66 & 41.55 & 41.64 & 39.13 & 30.56 & 40.28 \\
  \bottomrule
  \end{tabular}
  \caption{Language-based model performance, averaged over all tasks.}
  \label{tab:language_model_evaluation}
\end{table*}

\paragraph{Topic Effects.}
Performance remains stable across research topics for all evaluated agents, with all models (\textit{OpenAI}, \textit{Gemini}, \textit{Doubao}, \textit{Qwen}, \textit{Grok}, \textit{Perplexity}) showing no significant topic-based variation (p $>$ 0.05).

\subsection{Source Article Leakage}

Since the goal is to indirectly compare human expert-authored research reports with those generated by the model, any situation where the model is exposed to a human expert article during the research process would constitute a form of answer leakage. To mitigate this risk, we included a blocked list in the prompt to prevent the model from accessing these source articles. Additionally, during the evaluation, we performed a secondary inspection of the generated reports. If a report referenced the source article and correctly answered a question, we excluded that from the score and recorded the leakage rate. The results are presented in Appendix~\ref{sec:source-leakage}.

As most of the models evaluated are closed-source, we can only prevent the model from viewing the source articles through the prompt as a suboptimal workaround. However, this approach cannot guarantee 100\% prevention of model access to the source articles, and it may potentially affect the model's actual performance. Therefore, for open-source projects and internal members of closed-source teams, we strongly recommend using more restrictive measures (e.g., blocking certain search tool results or limiting searchable URLs) when generating evaluation reports to fully prevent model access to the source articles, ensuring more accurate evaluation results. The specific list of source articles is provided in the appendix. Each task in the public dataset is accompanied by the corresponding source article title, authors, and URLs.

\subsection{Future Directions}

In the previous subsection, we included the requirement in the Presentation dimension: Considering the user's cognitive level and background knowledge. However, this is a particularly challenging task. In our benchmark, this aspect has not been fully realized. Recent works have made initial efforts in this direction, such as LiveResearchBench, which introduces the concept of target\_audience, allowing the model to generate reports tailored to the user's background. However, this remains a very preliminary attempt. Simply adding one or two sentences about the target\_audience in the prompt does not effectively help the model understand the user's cognitive level and background knowledge. One cannot assume that an undergraduate student knows nothing and needs detailed explanations of every concept, nor can one assume that an experienced professor understands everything and requires no background context. Therefore, the true realization of user-adaptive presentation remains an open research direction, both for deep research agents and evaluation benchmarks. Achieving this would require collaborative efforts from research areas such as Agent Memory and User Modeling.

\section{Conclusion}
In this work, we have presented Deep Research Bench II, a comprehensive benchmark designed to evaluate the capabilities of deep research agents. By focusing on real-world user needs and deconstructing research tasks into three core dimensions---Information Recall, Analysis, and Presentation---we have created a more robust and realistic method for evaluating LLM-based research agents. Our approach incorporates expert-authored articles and verifiable rubrics, ensuring that the evaluation process aligns with human expert expectations and provides a reliable assessment of model performance. The results highlight the strengths and weaknesses of leading deep research models, offering valuable insights into areas where improvements are needed, such as reasoning and the integration of retrieved data. Additionally, we identify the challenges of user-adaptive presentations, which remains a promising avenue for future research. Deep Research Bench II sets a new standard for the evaluation of deep research agents, and we believe it will drive future advancements in this field by providing a more accurate measure of model capabilities and facilitating the development of more effective research tools.

\section{Limitations}

\paragraph{Source Article Leakage.} In our own experiments, we encountered limitations in controlling search results from closed-source models. We attempted to prevent source articles from being accessed by the agent by using prompt-based restrictions. However, the final statistics revealed that this method could not fully prevent leakage of source articles. The reliance on prompt-based restrictions may influence the performance of the deep research agent, and if the leakage rate is too high, it could lead to discrepancies between the reported scores and the agent's actual performance, thereby diminishing the relevance and accuracy of the evaluation results. To address this, a potential and strongly recommended solution is for the development team of deep research agents to directly restrict access to these articles at the search engine tool level, effectively eliminating the possibility of leakage and ensuring a more accurate evaluation of the system’s performance.

\paragraph{Human Annotations.} Although we invited human experts to annotate the benchmark and implemented multiple rounds of review to ensure the quality of annotations, the subjective judgment of human annotators could still introduce bias, affecting the final results of the benchmark. Even research reports authored by human experts are unlikely to satisfy all reviewers in a single pass, highlighting the inherently challenging nature of evaluating deep research agents. As such, the evaluation process—especially within the context of deep research—remains an ongoing, long-tail issue that requires continuous community efforts for refinement. To address this, we welcome feedback and suggestions from the broader community and domain experts to improve both the tasks and rubrics, ensuring that the benchmark evolves to better reflect human preferences.

\paragraph{Limitations in the Presentation Dimension.} As discussed in the "Future Directions" section, the current presentation dimension primarily assesses the formatting and layout of reports, but it does not yet account for how the model can personalize the presentation based on the user's knowledge background and preferences. Enhancing this aspect requires integration with advancements in agent memory, making it a key area for future improvements. We consider this to be one of the central directions for further development of the benchmark.

\section{Potential Risks}

\paragraph{Intellectual Property Concerns.} In our work, we directly incorporated several human-expert articles that were explicitly licensed under CC-by-4.0 or CC-BY-4.0-NC licenses, allowing non-commercial use and adaptation. Accordingly, our benchmark follows the same licensing terms, permitting free adaptation and usage under non-commercial conditions. However, there remains a potential risk that commercial entities could utilize this benchmark as training data, which could indirectly infringe upon the intellectual property rights of the original authors. While we have taken measures to ensure compliance with open-access guidelines, this risk persists and must be carefully monitored.

\section{Acknowledgements}

We would like to express our gratitude to all the annotating experts who contributed to our benchmark. Their dedicated work and professional input were invaluable to the completion of this project. Without their support, this work would not have been possible.

We also extend our sincere thanks to the authors of the source articles used in our benchmark. Your commitment to open access is truly appreciated. If any author prefers not to have their work included in our benchmark, we fully respect your decision. Please feel free to contact us, and we will promptly remove the article from our dataset.

We are also deeply thankful to all the peers and experts who have provided suggestions and assistance during the development of this work and after its publication. We thank Bei Zheng for helping us create the figures. Your contributions are helping us to continuously improve the benchmark. We encourage experts in related fields to continue offering feedback for future revisions of the benchmark, so that we can collaboratively address this long-tail challenge and contribute to the ongoing evolution of deep research agents.

\bibliographystyle{plainnat}
\bibliography{main}

@misc{gou2025mind2web2evaluatingagentic,
      title={Mind2Web 2: Evaluating Agentic Search with Agent-as-a-Judge}, 
      author={Boyu Gou and Zanming Huang and Yuting Ning and Yu Gu and Michael Lin and Weijian Qi and Andrei Kopanev and Botao Yu and Bernal Jiménez Gutiérrez and Yiheng Shu and Chan Hee Song and Jiaman Wu and Shijie Chen and Hanane Nour Moussa and Tianshu Zhang and Jian Xie and Yifei Li and Tianci Xue and Zeyi Liao and Kai Zhang and Boyuan Zheng and Zhaowei Cai and Viktor Rozgic and Morteza Ziyadi and Huan Sun and Yu Su},
      year={2025},
      eprint={2506.21506},
      archivePrefix={arXiv},
      primaryClass={cs.AI},
      url={https://arxiv.org/abs/2506.21506}, 
}

@misc{haas2025build-gemini-deep-research,
  author       = {Google},
  title        = {Build with Gemini Deep Research},
  year         = {2025},
  month        = dec,
  day          = {11},
  publisher    = {Google},
  note         = {The Keyword (Google Blog)},
  url          = {https://blog.google/technology/developers/deep-research-agent-gemini-api/},
  urldate      = {2026-01-01}
}

@misc{bytedance_doubao_chat,
  title        = {Doubao Chat},
  author       = {{ByteDance}},
  howpublished = {\url{https://www.doubao.com/chat/}},
  note         = {Accessed: 2025-12-19}
}

@misc{xai_grok,
  title        = {Grok},
  author       = {{xAI}},
  howpublished = {\url{https://grok.com/}},
  note         = {Accessed: 2025-12-19}
}

@misc{wang2025evaluatingllmsmultipleproblems,
      title={Evaluating LLMs with Multiple Problems at once}, 
      author={Zhengxiang Wang and Jordan Kodner and Owen Rambow},
      year={2025},
      eprint={2406.10786},
      archivePrefix={arXiv},
      primaryClass={cs.AI},
      url={https://arxiv.org/abs/2406.10786}, 
}

@misc{tongyidr,
  author={Tongyi DeepResearch Team},
  title={Tongyi DeepResearch: A New Era of Open-Source AI Researchers},
  year={2025},
  howpublished={\url{https://github.com/Alibaba-NLP/DeepResearch}}
}

@misc{OpenAI_DeepResearch_2025,
  author       = {OpenAI},
  title        = {Introducing Deep Research},
  howpublished = {\url{https://openai.com/index/introducing-deep-research/}},
  year         = {2025},
  note         = {Accessed: 2025‑11‑20},
}

@misc{Google_Gemini_DeepResearch_2024,
  author       = {Google},
  title        = {Try Deep Research and our new experimental model in Gemini},
  howpublished = {\url{https://blog.google/products/gemini/google-gemini-deep-research/}},
  month        = dec,
  year         = {2024},
  note         = {Accessed: 2025‑11‑20},
}

@misc{Perplexity_DeepResearch_2025,
  author       = {Perplexity AI},
  title        = {Introducing Deep Research},
  howpublished = {\url{https://www.perplexity.ai/hub/blog/introducing-perplexity-deep-research}},
  year         = {2025},
  note         = {Accessed: 2025‑11‑20},
}

@misc{Alibaba_Qwen_DeepResearch_2025,
  author       = {Alibaba Cloud Blog},
  title        = {Qwen DeepResearch: When Inspiration Becomes Its Own Reason},
  howpublished = {\url{https://www.alibabacloud.com/blog/qwen-deepresearch-when-inspiration-becomes-its-own-reason_602676}},
  year         = {2025},
  note         = {Accessed: 2025‑11‑20},
}

@article{Farquhar2024DetectingHL,
  author = {Farquhar, S. and Kossen, J. and Kuhn, L. and others},
  title = {Detecting hallucinations in large language models using semantic entropy},
  journal = {Nature},
  volume = {630},
  pages = {625--630},
  year = {2024},
  doi = {10.1038/s41586-024-07421-0},
  url = {https://doi.org/10.1038/s41586-024-07421-0},
  issue = {June 2024},
  received = {17 July 2023},
  accepted = {12 April 2024},
  published = {19 June 2024},
}

@misc{liu2025longcontexthallucinationdetection,
      title={Towards Long Context Hallucination Detection}, 
      author={Siyi Liu and Kishaloy Halder and Zheng Qi and Wei Xiao and Nikolaos Pappas and Phu Mon Htut and Neha Anna John and Yassine Benajiba and Dan Roth},
      year={2025},
      eprint={2504.19457},
      archivePrefix={arXiv},
      primaryClass={cs.CL},
      url={https://arxiv.org/abs/2504.19457}, 
}

@misc{simhi2025trustmeimwrong,
      title={Trust Me, I'm Wrong: LLMs Hallucinate with Certainty Despite Knowing the Answer}, 
      author={Adi Simhi and Itay Itzhak and Fazl Barez and Gabriel Stanovsky and Yonatan Belinkov},
      year={2025},
      eprint={2502.12964},
      archivePrefix={arXiv},
      primaryClass={cs.CL},
      url={https://arxiv.org/abs/2502.12964}, 
}

@article{Huang_2025,
   title={A Survey on Hallucination in Large Language Models: Principles, Taxonomy, Challenges, and Open Questions},
   volume={43},
   ISSN={1558-2868},
   url={http://dx.doi.org/10.1145/3703155},
   DOI={10.1145/3703155},
   number={2},
   journal={ACM Transactions on Information Systems},
   publisher={Association for Computing Machinery (ACM)},
   author={Huang, Lei and Yu, Weijiang and Ma, Weitao and Zhong, Weihong and Feng, Zhangyin and Wang, Haotian and Chen, Qianglong and Peng, Weihua and Feng, Xiaocheng and Qin, Bing and Liu, Ting},
   year={2025},
   month=jan, pages={1–55} }

@misc{huynh2025largelanguagemodelscode,
      title={Large Language Models for Code Generation: A Comprehensive Survey of Challenges, Techniques, Evaluation, and Applications}, 
      author={Nam Huynh and Beiyu Lin},
      year={2025},
      eprint={2503.01245},
      archivePrefix={arXiv},
      primaryClass={cs.SE},
      url={https://arxiv.org/abs/2503.01245}, 
}

@article{Singhal2023LargeLM,
  author = {Singhal, K. and Azizi, S. and Tu, T. and others},
  title = {Large language models encode clinical knowledge},
  journal = {Nature},
  volume = {620},
  pages = {172--180},
  year = {2023},
  doi = {10.1038/s41586-023-06291-2},
  url = {https://doi.org/10.1038/s41586-023-06291-2},
  issue = {July 2023},
  received = {25 January 2023},
  accepted = {05 June 2023},
  published = {12 July 2023},
}

@misc{pathak2025rubricneedenhancingllmbased,
      title={Rubric Is All You Need: Enhancing LLM-based Code Evaluation With Question-Specific Rubrics}, 
      author={Aditya Pathak and Rachit Gandhi and Vaibhav Uttam and Arnav Ramamoorthy and Pratyush Ghosh and Aaryan Raj Jindal and Shreyash Verma and Aditya Mittal and Aashna Ased and Chirag Khatri and Yashwanth Nakka and Devansh and Jagat Sesh Challa and Dhruv Kumar},
      year={2025},
      eprint={2503.23989},
      archivePrefix={arXiv},
      primaryClass={cs.SE},
      doi={https://doi.org/10.1145/3702652.3744220},
      url={https://arxiv.org/abs/2503.23989}, 
}

@inproceedings{Hashemi_2024,
   title={LLM-Rubric: A Multidimensional, Calibrated Approach to Automated Evaluation of Natural Language Texts},
   url={http://dx.doi.org/10.18653/v1/2024.acl-long.745},
   DOI={10.18653/v1/2024.acl-long.745},
   booktitle={Proceedings of the 62nd Annual Meeting of the Association for Computational Linguistics (Volume 1: Long Papers)},
   publisher={Association for Computational Linguistics},
   author={Hashemi, Helia and Eisner, Jason and Rosset, Corby and Van Durme, Benjamin and Kedzie, Chris},
   year={2024},
   pages={13806–13834} }

@misc{aroraHealthBenchEvaluatingLarge2025,
  title = {{{HealthBench}}: {{Evaluating Large Language Models Towards Improved Human Health}}},
  shorttitle = {{{HealthBench}}},
  author = {Arora, Rahul K. and Wei, Jason and Hicks, Rebecca Soskin and Bowman, Preston and {Qui{\~n}onero-Candela}, Joaquin and Tsimpourlas, Foivos and Sharman, Michael and Shah, Meghan and Vallone, Andrea and Beutel, Alex and Heidecke, Johannes and Singhal, Karan},
  year = 2025,
  month = may,
  number = {arXiv:2505.08775},
  eprint = {2505.08775},
  primaryclass = {cs},
  publisher = {arXiv},
  doi = {10.48550/arXiv.2505.08775},
  urldate = {2025-08-07},
  archiveprefix = {arXiv},
  langid = {english}
}

@misc{lin2024wildbenchbenchmarkingllmschallenging,
      title={WildBench: Benchmarking LLMs with Challenging Tasks from Real Users in the Wild}, 
      author={Bill Yuchen Lin and Yuntian Deng and Khyathi Chandu and Faeze Brahman and Abhilasha Ravichander and Valentina Pyatkin and Nouha Dziri and Ronan Le Bras and Yejin Choi},
      year={2024},
      eprint={2406.04770},
      archivePrefix={arXiv},
      primaryClass={cs.CL},
      url={https://arxiv.org/abs/2406.04770}, 
}

@misc{sirdeshmukh2025multichallengerealisticmultiturnconversation,
      title={MultiChallenge: A Realistic Multi-Turn Conversation Evaluation Benchmark Challenging to Frontier LLMs}, 
      author={Ved Sirdeshmukh and Kaustubh Deshpande and Johannes Mols and Lifeng Jin and Ed-Yeremai Cardona and Dean Lee and Jeremy Kritz and Willow Primack and Summer Yue and Chen Xing},
      year={2025},
      eprint={2501.17399},
      archivePrefix={arXiv},
      primaryClass={cs.CL},
      url={https://arxiv.org/abs/2501.17399}, 
}

@misc{starace2025paperbenchevaluatingaisability,
      title={PaperBench: Evaluating AI's Ability to Replicate AI Research}, 
      author={Giulio Starace and Oliver Jaffe and Dane Sherburn and James Aung and Jun Shern Chan and Leon Maksin and Rachel Dias and Evan Mays and Benjamin Kinsella and Wyatt Thompson and Johannes Heidecke and Amelia Glaese and Tejal Patwardhan},
      year={2025},
      eprint={2504.01848},
      archivePrefix={arXiv},
      primaryClass={cs.AI},
      url={https://arxiv.org/abs/2504.01848}, 
}

@misc{li2025automatedcreativityevaluationlarge,
      title={Automated Creativity Evaluation for Large Language Models: A Reference-Based Approach}, 
      author={Ruizhe Li and Chiwei Zhu and Benfeng Xu and Xiaorui Wang and Zhendong Mao},
      year={2025},
      eprint={2504.15784},
      archivePrefix={arXiv},
      primaryClass={cs.CL},
      url={https://arxiv.org/abs/2504.15784}, 
}

@misc{chakrabarty2024artartificelargelanguage,
      title={Art or Artifice? Large Language Models and the False Promise of Creativity}, 
      author={Tuhin Chakrabarty and Philippe Laban and Divyansh Agarwal and Smaranda Muresan and Chien-Sheng Wu},
      year={2024},
      eprint={2309.14556},
      archivePrefix={arXiv},
      primaryClass={cs.CL},
      url={https://arxiv.org/abs/2309.14556}, 
}

@misc{marioriyad2025silentjudgeunacknowledgedshortcut,
      title={The Silent Judge: Unacknowledged Shortcut Bias in LLM-as-a-Judge}, 
      author={Arash Marioriyad and Mohammad Hossein Rohban and Mahdieh Soleymani Baghshah},
      year={2025},
      eprint={2509.26072},
      archivePrefix={arXiv},
      primaryClass={cs.CL},
      url={https://arxiv.org/abs/2509.26072}, 
}

@misc{thakur2025judgingjudgesevaluatingalignment,
      title={Judging the Judges: Evaluating Alignment and Vulnerabilities in LLMs-as-Judges}, 
      author={Aman Singh Thakur and Kartik Choudhary and Venkat Srinik Ramayapally and Sankaran Vaidyanathan and Dieuwke Hupkes},
      year={2025},
      eprint={2406.12624},
      archivePrefix={arXiv},
      primaryClass={cs.CL},
      url={https://arxiv.org/abs/2406.12624}, 
}

@misc{bavaresco2025llmsinsteadhumanjudges,
      title={LLMs instead of Human Judges? A Large Scale Empirical Study across 20 NLP Evaluation Tasks}, 
      author={Anna Bavaresco and Raffaella Bernardi and Leonardo Bertolazzi and Desmond Elliott and Raquel Fernández and Albert Gatt and Esam Ghaleb and Mario Giulianelli and Michael Hanna and Alexander Koller and André F. T. Martins and Philipp Mondorf and Vera Neplenbroek and Sandro Pezzelle and Barbara Plank and David Schlangen and Alessandro Suglia and Aditya K Surikuchi and Ece Takmaz and Alberto Testoni},
      year={2025},
      eprint={2406.18403},
      archivePrefix={arXiv},
      primaryClass={cs.CL},
      url={https://arxiv.org/abs/2406.18403}, 
}

@misc{gu2025surveyllmasajudge,
      title={A Survey on LLM-as-a-Judge}, 
      author={Jiawei Gu and Xuhui Jiang and Zhichao Shi and Hexiang Tan and Xuehao Zhai and Chengjin Xu and Wei Li and Yinghan Shen and Shengjie Ma and Honghao Liu and Saizhuo Wang and Kun Zhang and Yuanzhuo Wang and Wen Gao and Lionel Ni and Jian Guo},
      year={2025},
      eprint={2411.15594},
      archivePrefix={arXiv},
      primaryClass={cs.CL},
      url={https://arxiv.org/abs/2411.15594}, 
}

@misc{zheng2023judgingllmasajudgemtbenchchatbot,
      title={Judging LLM-as-a-Judge with MT-Bench and Chatbot Arena}, 
      author={Lianmin Zheng and Wei-Lin Chiang and Ying Sheng and Siyuan Zhuang and Zhanghao Wu and Yonghao Zhuang and Zi Lin and Zhuohan Li and Dacheng Li and Eric P. Xing and Hao Zhang and Joseph E. Gonzalez and Ion Stoica},
      year={2023},
      eprint={2306.05685},
      archivePrefix={arXiv},
      primaryClass={cs.CL},
      url={https://arxiv.org/abs/2306.05685}, 
}

@misc{sharmaResearchRubricsBenchmarkPrompts2025,
  title = {{{ResearchRubrics}}: {{A Benchmark}} of {{Prompts}} and {{Rubrics For Evaluating Deep Research Agents}}},
  shorttitle = {{{ResearchRubrics}}},
  author = {Sharma, Manasi and Zhang, Chen Bo Calvin and Bandi, Chaithanya and Wang, Clinton and Aich, Ankit and Nghiem, Huy and Rabbani, Tahseen and Htet, Ye and Jang, Brian and Basu, Sumana and Balwani, Aishwarya and Peskoff, Denis and Ayestaran, Marcos and Hendryx, Sean M. and Kenstler, Brad and Liu, Bing},
  year = 2025,
  month = nov,
  number = {arXiv:2511.07685},
  eprint = {2511.07685},
  primaryclass = {cs},
  publisher = {arXiv},
  doi = {10.48550/arXiv.2511.07685},
  urldate = {2025-11-18},
  archiveprefix = {arXiv},
  langid = {english}
}

@misc{wangLiveResearchBenchLiveBenchmark2025,
  title = {{{LiveResearchBench}}: {{A Live Benchmark}} for {{User-Centric Deep Research}} in the {{Wild}}},
  shorttitle = {{{LiveResearchBench}}},
  author = {Wang, Jiayu and Ming, Yifei and Dulepet, Riya and Chen, Qinglin and Xu, Austin and Ke, Zixuan and Sala, Frederic and Albarghouthi, Aws and Xiong, Caiming and Joty, Shafiq},
  year = 2025,
  month = nov,
  number = {arXiv:2510.14240},
  eprint = {2510.14240},
  primaryclass = {cs},
  publisher = {arXiv},
  doi = {10.48550/arXiv.2510.14240},
  urldate = {2025-11-18},
  archiveprefix = {arXiv},
  langid = {english}
}

@misc{brown2020languagemodelsfewshotlearners,
      title={Language Models are Few-Shot Learners}, 
      author={Tom B. Brown and Benjamin Mann and Nick Ryder and Melanie Subbiah and Jared Kaplan and Prafulla Dhariwal and Arvind Neelakantan and Pranav Shyam and Girish Sastry and Amanda Askell and Sandhini Agarwal and Ariel Herbert-Voss and Gretchen Krueger and Tom Henighan and Rewon Child and Aditya Ramesh and Daniel M. Ziegler and Jeffrey Wu and Clemens Winter and Christopher Hesse and Mark Chen and Eric Sigler and Mateusz Litwin and Scott Gray and Benjamin Chess and Jack Clark and Christopher Berner and Sam McCandlish and Alec Radford and Ilya Sutskever and Dario Amodei},
      year={2020},
      eprint={2005.14165},
      archivePrefix={arXiv},
      primaryClass={cs.CL},
      url={https://arxiv.org/abs/2005.14165}, 
}

@misc{andreas2022languagemodelsagentmodels,
      title={Language Models as Agent Models}, 
      author={Jacob Andreas},
      year={2022},
      eprint={2212.01681},
      archivePrefix={arXiv},
      primaryClass={cs.CL},
      url={https://arxiv.org/abs/2212.01681}, 
}

@misc{nakano2022webgptbrowserassistedquestionansweringhuman,
      title={WebGPT: Browser-assisted question-answering with human feedback}, 
      author={Reiichiro Nakano and Jacob Hilton and Suchir Balaji and Jeff Wu and Long Ouyang and Christina Kim and Christopher Hesse and Shantanu Jain and Vineet Kosaraju and William Saunders and Xu Jiang and Karl Cobbe and Tyna Eloundou and Gretchen Krueger and Kevin Button and Matthew Knight and Benjamin Chess and John Schulman},
      year={2022},
      eprint={2112.09332},
      archivePrefix={arXiv},
      primaryClass={cs.CL},
      url={https://arxiv.org/abs/2112.09332}, 
}

@article{White2024Advancing,
  author       = {Ryen W. White},
  title        = {Advancing the Search Frontier with AI Agents},
  journal      = {Communications of the ACM},
  volume       = {67},
  number       = {9},
  pages        = {54--65},
  year         = {2024},
  doi          = {10.1145/3655615}
}

@misc{wei2025browsecompsimplechallengingbenchmark,
      title={BrowseComp: A Simple Yet Challenging Benchmark for Browsing Agents}, 
      author={Jason Wei and Zhiqing Sun and Spencer Papay and Scott McKinney and Jeffrey Han and Isa Fulford and Hyung Won Chung and Alex Tachard Passos and William Fedus and Amelia Glaese},
      year={2025},
      eprint={2504.12516},
      archivePrefix={arXiv},
      primaryClass={cs.CL},
      url={https://arxiv.org/abs/2504.12516}, 
}

@misc{chen2025xbenchtrackingagentsproductivity,
      title={xbench: Tracking Agents Productivity Scaling with Profession-Aligned Real-World Evaluations}, 
      author={Kaiyuan Chen and Yixin Ren and Yang Liu and Xiaobo Hu and Haotong Tian and Tianbao Xie and Fangfu Liu and Haoye Zhang and Hongzhang Liu and Yuan Gong and Chen Sun and Han Hou and Hui Yang and James Pan and Jianan Lou and Jiayi Mao and Jizheng Liu and Jinpeng Li and Kangyi Liu and Kenkun Liu and Rui Wang and Run Li and Tong Niu and Wenlong Zhang and Wenqi Yan and Xuanzheng Wang and Yuchen Zhang and Yi-Hsin Hung and Yuan Jiang and Zexuan Liu and Zihan Yin and Zijian Ma and Zhiwen Mo},
      year={2025},
      eprint={2506.13651},
      archivePrefix={arXiv},
      primaryClass={cs.LG},
      url={https://arxiv.org/abs/2506.13651}, 
}

@misc{futuresearch2025deepresearchbenchevaluating,
      title={Deep Research Bench: Evaluating AI Web Research Agents}, 
      author={FutureSearch and : and Nikos I. Bosse and Jon Evans and Robert G. Gambee and Daniel Hnyk and Peter Mühlbacher and Lawrence Phillips and Dan Schwarz and Jack Wildman},
      year={2025},
      eprint={2506.06287},
      archivePrefix={arXiv},
      primaryClass={cs.AI},
      url={https://arxiv.org/abs/2506.06287}, 
}

@misc{du2025deepresearchbenchcomprehensivebenchmark,
      title={DeepResearch Bench: A Comprehensive Benchmark for Deep Research Agents}, 
      author={Mingxuan Du and Benfeng Xu and Chiwei Zhu and Xiaorui Wang and Zhendong Mao},
      year={2025},
      eprint={2506.11763},
      archivePrefix={arXiv},
      primaryClass={cs.CL},
      url={https://arxiv.org/abs/2506.11763}, 
}

@misc{wong2025widesearchbenchmarkingagenticbroad,
      title={WideSearch: Benchmarking Agentic Broad Info-Seeking}, 
      author={Ryan Wong and Jiawei Wang and Junjie Zhao and Li Chen and Yan Gao and Long Zhang and Xuan Zhou and Zuo Wang and Kai Xiang and Ge Zhang and Wenhao Huang and Yang Wang and Ke Wang},
      year={2025},
      eprint={2508.07999},
      archivePrefix={arXiv},
      primaryClass={cs.CL},
      url={https://arxiv.org/abs/2508.07999}, 
}

@misc{xu2025researcherbenchevaluatingdeepai,
      title={ResearcherBench: Evaluating Deep AI Research Systems on the Frontiers of Scientific Inquiry}, 
      author={Tianze Xu and Pengrui Lu and Lyumanshan Ye and Xiangkun Hu and Pengfei Liu},
      year={2025},
      eprint={2507.16280},
      archivePrefix={arXiv},
      primaryClass={cs.AI},
      url={https://arxiv.org/abs/2507.16280}, 
}

@misc{fan2025understandingdeepresearchreports,
      title={Understanding DeepResearch via Reports}, 
      author={Tianyu Fan and Xinyao Niu and Yuxiang Zheng and Fengji Zhang and Chengen Huang and Bei Chen and Junyang Lin and Chao Huang},
      year={2025},
      eprint={2510.07861},
      archivePrefix={arXiv},
      primaryClass={cs.AI},
      url={https://arxiv.org/abs/2510.07861}, 
}

@misc{gemini_deep_research,
  title        = {Gemini Deep Research — your personal research assistant},
  author       = {{Google}},
  year         = {2024},
  howpublished = {\url{https://gemini.google/overview/deep-research/}},
  note         = {Accessed 2025-11-10}
}

@misc{openai_deep_research,
  title        = {Introducing Deep Research},
  author       = {{OpenAI}},
  year         = {2025},
  howpublished = {\url{https://openai.com/index/introducing-deep-research/}},
  note         = {Accessed 2025-11-10}
}

\clearpage

\onecolumn
\appendix

\section{Benchmark Statistics}
\subsection{Task Topic and Language Statistics}

\begin{table*}[h]
\centering
\begin{tabular}{lccc}
\toprule
\textbf{Topic} & \textbf{EN} & \textbf{ZH} & \textbf{Total} \\
\midrule
Science \& Technology & 12 & 11 & 23 \\
Finance \& Business & 7 & 8 & 15 \\
Software Development & 6 & 7 & 13 \\
Health & 6 & 6 & 12 \\
Education \& Jobs & 5 & 5 & 10 \\
Hardware & 3 & 3 & 6 \\
History & 3 & 3 & 6 \\
Literature & 3 & 3 & 6 \\
Industrial & 3 & 2 & 5 \\
Art \& Design & 2 & 2 & 4 \\
Religion & 2 & 2 & 4 \\
Food \& Dining & 2 & 1 & 3 \\
Home \& Hobbies & 2 & 1 & 3 \\
Social Life & 1 & 2 & 3 \\
Sports \& Fitness & 2 & 1 & 3 \\
Transportation & 1 & 2 & 3 \\
Travel & 1 & 2 & 3 \\
Crime \& Law & 1 & 1 & 2 \\
Entertainment & 1 & 1 & 2 \\
Fashion \& Beauty & 1 & 1 & 2 \\
Games & 1 & 1 & 2 \\
Software & 1 & 1 & 2 \\
\midrule
\textbf{Total} & \textbf{66} & \textbf{66} & \textbf{132} \\
\bottomrule
\end{tabular}
\caption{Task topic and language distribution for DRB-v2, showing the number of tasks in English (EN), Chinese (ZH), and total count across different topic categories.}
\label{tab:task-topic-language-distribution}
\end{table*}

\subsection{Rubric Statistics by Dimension}

\begin{figure*}[h]
\centering
\includegraphics[width=0.8\textwidth]{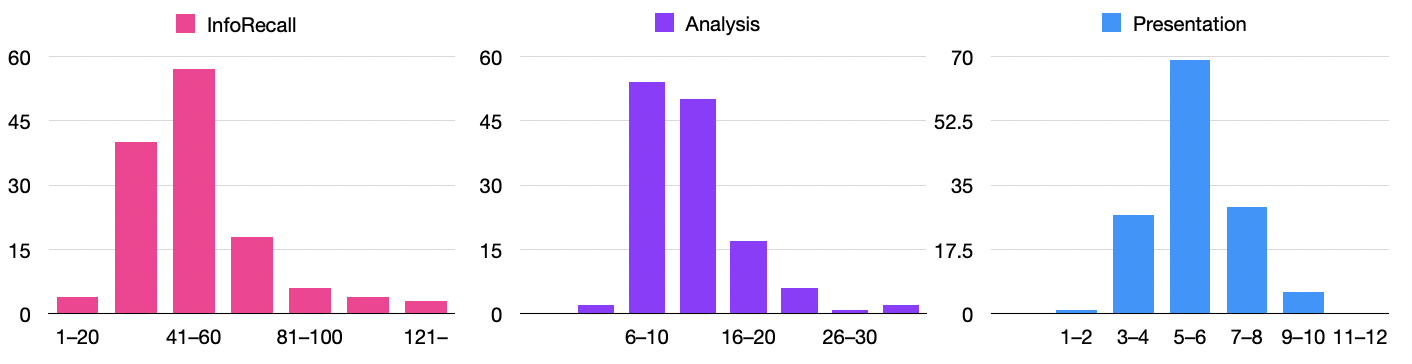}
\caption{Frequency distribution of rubric counts per task across the three evaluation dimensions: InfoRecall, Analysis, and Presentation. The distributions show the concentration of rubric counts within specific ranges for each dimension.}
\label{fig:rubric-distribution}
\end{figure*}

We analyze the distribution of rubrics across the three evaluation dimensions. On average, each task contains 52.902 rubrics for \textit{InfoRecall}, 12.773 rubrics for \textit{Analysis}, and 5.652 rubrics for \textit{Presentation}. Figure~\ref{fig:rubric-distribution} shows the frequency distribution of rubric counts across all tasks for each dimension.

\section{Detailed Result}

\begin{table}[H]
  \centering
  \small
  \begin{tabular}{lcccccccc}
  \toprule
  \textbf{Category} & \textbf{OpenAI} & \textbf{Gemini} & \textbf{Doubao} & \textbf{Qwen} & \textbf{Grok} & \textbf{Perplexity} & \textbf{Tongyi} & \textbf{Avg.} \\
  \midrule
  Art \& Design      & 0.5001 & 0.4887 & 0.4085 & 0.4675 & 0.4846 & 0.4292 & 0.2896 & 0.4383 \\
  Crime \& Law       & 0.5472 & 0.4667 & 0.5556 & 0.2222 & 0.4819 & 0.4861 & 0.2278 & 0.4268 \\
  Education \& Jobs  & 0.3460 & 0.3676 & 0.3540 & 0.4063 & 0.3352 & 0.3304 & 0.2682 & 0.3440 \\
  Entertainment      & 0.3647 & 0.4361 & 0.3678 & 0.3148 & 0.3375 & 0.2957 & 0.2017 & 0.3312 \\
  Fashion \& Beauty  & 0.4606 & 0.2524 & 0.2972 & 0.7193 & 0.2466 & 0.2042 & 0.2031 & 0.3405 \\
  Finance \& Business& 0.5336 & 0.5099 & 0.4514 & 0.4132 & 0.4741 & 0.4145 & 0.3560 & 0.4504 \\
  Food \& Dining     & 0.3283 & 0.4419 & 0.2782 & 0.4174 & 0.3610 & 0.3927 & 0.2714 & 0.3558 \\
  Games              & 0.3993 & 0.3887 & 0.3352 & 0.3569 & 0.3402 & 0.2766 & 0.3006 & 0.3425 \\
  Hardware           & 0.3652 & 0.4147 & 0.3096 & 0.3402 & 0.3536 & 0.3494 & 0.2466 & 0.3399 \\
  Health             & 0.5249 & 0.4493 & 0.4711 & 0.4308 & 0.4573 & 0.4635 & 0.3295 & 0.4466 \\
  History            & 0.4554 & 0.3914 & 0.4786 & 0.3552 & 0.4030 & 0.4183 & 0.2886 & 0.3986 \\
  Home \& Hobbies    & 0.5712 & 0.5235 & 0.4681 & 0.5194 & 0.4561 & 0.5311 & 0.3646 & 0.4906 \\
  Industrial         & 0.3793 & 0.3122 & 0.2978 & 0.2871 & 0.3262 & 0.3320 & 0.2427 & 0.3110 \\
  Literature         & 0.4868 & 0.4148 & 0.4664 & 0.4613 & 0.4066 & 0.3810 & 0.3124 & 0.4185 \\
  Religion           & 0.4255 & 0.3798 & 0.4703 & 0.4472 & 0.4356 & 0.3587 & 0.3024 & 0.4028 \\
  Science \& Tech    & 0.4536 & 0.4062 & 0.4214 & 0.3666 & 0.3625 & 0.3727 & 0.3085 & 0.3845 \\
  Social Life        & 0.3659 & 0.2996 & 0.3496 & 0.1963 & 0.3016 & 0.2866 & 0.2116 & 0.2873 \\
  Software           & 0.6679 & 0.5167 & 0.4821 & 0.4567 & 0.4421 & 0.5523 & 0.4757 & 0.5134 \\
  Software Dev.      & 0.4561 & 0.4504 & 0.4264 & 0.4218 & 0.4143 & 0.3819 & 0.3318 & 0.4118 \\
  Sports \& Fitness  & 0.2563 & 0.3004 & 0.2731 & 0.2597 & 0.2640 & 0.3216 & 0.2560 & 0.2759 \\
  Transportation     & 0.3840 & 0.4026 & 0.4313 & 0.2901 & 0.2744 & 0.3692 & 0.1880 & 0.3342 \\
  Travel             & 0.5568 & 0.3714 & 0.4204 & 0.3798 & 0.4060 & 0.4231 & 0.1939 & 0.3931 \\
  \bottomrule
  \end{tabular}
  \caption{Model performance across thematic categories.}
  \label{tab:model_evaluation_themes}
\end{table}

\section{Source Article List}
\label{sec:source-article-list}

\begin{longtable}{cp{8cm}p{5cm}}
  \caption{Blocked Articles} \label{tab:blocked_articles} \\
  \toprule
  \textbf{Idx} & \textbf{Title} & \textbf{Authors} \\
  \midrule
  \endfirsthead
  
  \multicolumn{3}{c}{\tablename\ \thetable\ -- \textit{Continued from previous page}} \\
  \toprule
  \textbf{Idx} & \textbf{Title} & \textbf{Authors} \\
  \midrule
  \endhead
  
  \midrule
  \multicolumn{3}{r}{\textit{Continued on next page}} \\
  \endfoot
  
  \bottomrule
  \endlastfoot
  1 & \href{https://www.mdpi.com/2073-445X/11/9/1529}{The Local Land Finance Transformation with the Synergy of Increment and Inventory: A Case Study in China} & Yuzhe Wu; Huiqiong Zhu; Sheng Zheng \\
  2 & \href{https://journals.plos.org/globalpublichealth/article?id=10.1371/journal.pgph.0002710}{South Asia's unprotected poor: A systematic review of why social protection programs fail to reach their potential} & Warda Javed; Zubia Mumtaz \\
  3 & \href{https://www.mdpi.com/1660-4601/19/24/16995}{Digital-Based Policy and Health Promotion Policy in Japan, the Republic of Korea, Singapore, and Thailand: A Scoping Review of Policy Paths to Healthy Aging} & Nadila Mulati; Myo Nyein Aung; Malcolm Field; Eun Woo Nam; Carol Ma Hok Ka; Saiyud Moolphate; Hocheol Lee; Yuki Goto; Nam Hae Kweun; Takumi Suda; Yuka Koyanagi; Yuiko Nagamine; Motoyuki Yuasa \\
  4 & \href{https://www.oecd.org/en/publications/protection-gaps-in-insurance-for-natural-hazards-and-retirement-savings-in-asia_294f044e-en.html}{Protection Gaps in Insurance for Natural Hazards and Retirement Savings in Asia} & OECD \\
  5 & \href{https://www.mdpi.com/1911-8074/14/11/551}{How Many Stocks Are Sufficient for Equity Portfolio Diversification? A Review of the Literature} & Azra Zaimovic; Adna Omanovic; Almira Arnaut-Berilo \\
  6 & \href{https://www.mdpi.com/2075-471X/12/6/91}{To Enhance the Credibility of the Green Bond Market through Regulating GBERs: The Case of China} & Xiayang Chen; Weiqiu Long \\
  7 & \href{https://doi.org/10.1108/PAP-08-2020-0037}{Sovereign wealth funds and corporate social responsibility: a comparison of Norway's Government Pension Fund Global and Abu Dhabi Fund for Development} & Sivakumar Velayutham; Rashedul Hasan \\
  8 & \href{https://journals.plos.org/plosone/article?id=10.1371/journal.pone.0282631}{Gold and silver as safe havens: A fractional integration and cointegration analysis} & Guglielmo Maria Caporale; Luis Alberiko Gil-Alana \\
  9 & \href{https://journals.plos.org/plosone/article?id=10.1371/journal.pone.0251752}{Is this time really different? Flight-to-safety and the COVID-19 crisis} & Celina Löwen; Bilal Kchouri; Thorsten Lehnert \\
  10 & \href{https://arxiv.org/abs/2503.01591}{The Role of Deep Learning in Financial Asset Management: A Systematic Review} & Pedro Dias Reis; Ana Paula Serra; João Gama \\
  11 & \href{https://www.mdpi.com/1099-4300/22/7/711}{A Review of Micro-Based Systemic Risk Research from Multiple Perspectives} & Xiao Bai; Huaping Sun; Shibao Lu; Farhad Taghizadeh-Hesary \\
  12 & \href{https://www.frontiersin.org/journals/artificial-intelligence/articles/10.3389/frai.2024.1371502/full}{Enhancing portfolio management using artificial intelligence: literature review} & Kristina Sutiene; Peter Schwendner; Ciprian Sipos; Luis Lorenzo; Miroslav Mirchev; Petre Lameski; Audrius Kabasinskas; Chemseddine Tidjani; Belma Ozturkkal; Jurgita Cerneviciene \\
  13 & \href{https://www.mdpi.com/2079-9292/14/1/153}{Autonomous Forklifts: State of the Art—Exploring Perception, Scanning Technologies and Functional Systems—A Comprehensive Review} & Muftah A Fraifer; Joseph Coleman; James Maguire; Petar Trslić; Gerard Dooly; Daniel Toal \\
  14 & \href{https://www.mdpi.com/2076-3387/14/10/238}{Artificial Intelligence in Auditing: A Conceptual Framework for Auditing Practices} & Diogo Leocádio; Luís Malheiro; João Reis \\
  15 & \href{https://www.mdpi.com/2227-9709/11/3/64}{Navigating Governmental Choices: A Comprehensive Review of Artificial Intelligence's Impact on Decision-Making} & Gustavo Caiza; Verónica Sanguña; Natalia Tusa; Violeta Masaquiza; Alexandra Ortiz; Marcelo V. Garcia \\
  16 & \href{https://www.techscience.com/cmc/v82n2/59511}{Machine Learning-Based Methods for Materials Inverse Design: A Review} & Yingli Liu; Yuting Cui; Haihe Zhou; Sheng Lei; Haibin Yuan; Tao Shen; Jiancheng Yin \\
  17 & \href{https://www.nature.com/articles/s41524-022-00734-6}{Recent advances and applications of deep learning methods in materials science} & Kamal Choudhary; Brian DeCost; Chi Chen; Anubhav Jain; Francesca Tavazza; Ryan Cohn; Cheol Woo Park; Alok Choudhary; Ankit Agrawal; Simon J. L. Billinge; Elizabeth Holm; Shyue Ping Ong; Chris Wolverton \\
  18 & \href{https://www.mdpi.com/2075-1729/11/8/857}{Horizontal Gene Transfers in Plants} & Emilie Aubin; Moaine El Baidouri; Olivier Panaud \\
  19 & \href{https://www.mdpi.com/1420-3049/28/2/695}{Effects and Influence of External Electric Fields on the Equilibrium Properties of Tautomeric Molecules} & Ivan Angelov; Lidia Zaharieva; Liudmil Antonov \\
  20 & \href{https://www.frontiersin.org/journals/physiology/articles/10.3389/fphys.2021.667000/full}{The Magnetic Compass of Birds: The Role of Cryptochrome} & Roswitha Wiltschko; Christine Nießner; Wolfgang Wiltschko \\
  21 & \href{https://link.springer.com/article/10.1140/epjs/s11734-022-00610-w}{Animal navigation: how animals use environmental factors to find their way} & Roswitha Wiltschko; Wolfgang Wiltschko \\
  22 & \href{https://www.mdpi.com/2071-1050/12/6/2387}{Life Cycle Cost Assessment of Electric Vehicles: A Review and Bibliometric Analysis} & Bamidele Victor Ayodele; Siti Indati Mustapa \\
  23 & \href{https://www.mdpi.com/1996-1944/17/1/239}{Technological Advances and Market Developments of Solid-State Batteries: A Review} & Felix Thomas; Lauren Mahdi; Julien Lemaire; Diogo M. F. Santos \\
  24 & \href{https://arxiv.org/abs/2408.03261}{A Comprehensive Review on Cislunar Expansion and Space Domain Awareness} & Brian Baker-McEvilly; Surabhi Bhadauria; David Canales; Carolin Frueh \\
  25 & \href{https://www.mdpi.com/1996-1944/15/6/2023}{Organic Compounds as Corrosion Inhibitors for Carbon Steel in HCl Solution: A Comprehensive Review} & Liangyuan Chen; Dongzhu Lu; Yanhu Zhang \\
  26 & \href{https://www.researchgate.net/publication/371675999_A_Review_of_Inorganic_Corrosion_Inhibitors_Types_Mechanisms_and_Applications}{A Review of Inorganic Corrosion Inhibitors: Types, Mechanisms, and Applications} & Ahmed A. Al-Amiery; Emad Yousif; Wan Nor Roslam Wan Isahak; Waleed Khalid Al-Azzawi \\
  27 & \href{https://www.mdpi.com/2410-3888/8/2/80}{Stock Assessment of Chub Mackerel (Scomber japonicus) in the Northwest Pacific Using a Multi-Model Approach} & Kai Cai; Richard Kindong; Qiuyun Ma; Siquan Tian \\
  28 & \href{https://www.mdpi.com/2073-4441/16/15/2129}{Comprehensive Review for Energy Recovery Technologies Used in Water Distribution Systems Considering Their Performance, Technical Challenges, and Economic Viability} & Admitos A. Bideris-Davos; Panagis N. Vovos \\
  29 & \href{https://arxiv.org/abs/2206.15383}{Integrated Photonic Platforms for Quantum Technology: A Review} & Rohit K Ramakrishnan; Aravinth Balaji Ravichandran; Arpita Mishra; Archana Kaushalram; Gopalkrishna Hegde; Srinivas Talabattula; Peter P Rohde \\
  30 & \href{https://eurradiolexp.springeropen.com/articles/10.1186/s41747-023-00413-1}{Empowering PET: harnessing deep learning for improved clinical insight} & Alessia Artesani; Alessandro Bruno; Fabrizia Gelardi; Arturo Chiti \\
  31 & \href{https://www.mdpi.com/1424-8220/21/4/1051}{State-of-the-Art Mobile Radiation Detection Systems for Different Scenarios} & Luís Marques; Alberto Vale; Pedro Vaz \\
  32 & \href{https://www.mdpi.com/2073-4352/10/11/973}{Lithium Niobate Single Crystals and Powders Reviewed—Part I} & Oswaldo Sánchez-Dena; Cesar David Fierro-Ruiz; Sergio David Villalobos-Mendoza; Diana María Carrillo Flores; José Trinidad Elizalde-Galindo; Rurik Farías \\
  33 & \href{https://arxiv.org/abs/1912.06938}{Quantum Simulators: Architectures and Opportunities} & Ehud Altman; Kenneth R. Brown; Giuseppe Carleo; Lincoln D. Carr; Eugene Demler; Cheng Chin; Brian DeMarco; Sophia E. Economou; Mark A. Eriksson; Kai-Mei C. Fu; Markus Greiner; Kaden R. A. Hazzard; Randall G. Hulet; Alicia J. Kollar; Benjamin L. Lev; Mikhail D. Lukin; Ruichao Ma; Xiao Mi; Shashank Misra; Christopher Monroe; Kater Murch; Zaira Nazario; Kang-Kuen Ni; Andrew C. Potter; Pedram Roushan; Mark Saffman; Monika Schleier-Smith; Irfan Siddiqi; Raymond Simmonds; Meenakshi Singh; I. B. Spielman; Kristan Temme; David S. Weiss; Jelena Vuckovic; Vladan Vuletic; Jun Ye; Martin Zwierlein \\
  34 & \href{https://www.mdpi.com/2079-9292/13/23/4785}{A Survey of Open-Source UAV Autopilots} & Nourdine Aliane \\
  35 & \href{https://www.mdpi.com/2227-7080/9/2/37}{A Review on Comparative Remarks, Performance Evaluation and Improvement Strategies of Quadrotor Controllers} & Rupal Roy; Maidul Islam; Nafiz Sadman; M. A. Parvez Mahmud; Kishor Datta Gupta; Md Manjurul Ahsan \\
  36 & \href{https://epjquantumtechnology.springeropen.com/articles/10.1140/epjqt/s40507-022-00150-1}{Towards European standards for quantum technologies} & Oskar van Deventer; Nicolas Spethmann; Marius Loeffler; Michele Amoretti; Rob van den Brink; Natalia Bruno; Paolo Comi; Noel Farrugia; Marco Gramegna; Andreas Jenet; Ben Kassenberg; Wojciech Kozlowski; Thomas Länger; Tobias Lindstrom; Vicente Martin; Niels Neumann; Homer Papadopoulos; Saverio Pascazio; Momtchil Peev; Richard Pitwon; M. Adriaan Rol; Paolo Traina; Pim Venderbosch; Frank K. Wilhelm-Mauch \\
  37 & \href{https://arxiv.org/abs/1904.10042}{Self-testing of quantum systems: a review} & Ivan Šupić; Joseph Bowles \\
  38 & \href{https://www.mdpi.com/1424-8220/25/5/1388}{A Review of Environmental Control Strategies and Models for Modern Agricultural Greenhouses} & Shuailiang Chen; Aolong Liu; Fei Tang; Pei Hou; Yanli Lu; Pei Yuan \\
  39 & \href{https://www.mdpi.com/1424-8220/24/12/3963}{Non-Contact Vision-Based Techniques of Vital Sign Monitoring: Systematic Review} & Linas Saikevičius; Vidas Raudonis; Gintaras Dervinis; Virginijus Baranauskas \\
  40 & \href{https://www.mdpi.com/1424-8220/25/6/1913}{A Survey on Secure WiFi Sensing Technology: Attacks and Defenses} & Xingyu Liu; Xin Meng; Hancong Duan; Ze Hu; Min Wang \\
  41 & \href{https://link.springer.com/article/10.1007/s10270-021-00964-0}{Modelling in low-code development: a multi-vocal systematic review} & Alessio Bucaioni; Antonio Cicchetti; Federico Ciccozzi \\
  42 & \href{https://www.mdpi.com/2076-3417/15/12/6481}{Democratizing Digital Transformation: A Multisector Study of Low-Code Adoption Patterns, Limitations, and Emerging Paradigms} & Zhengwu Shi; Junyu Dong; Yanhai Gan \\
  43 & \href{https://www.mdpi.com/2227-7390/11/10/2234}{A Survey on Population-Based Deep Reinforcement Learning} & Weifan Long; Taixian Hou; Xiaoyi Wei; Shichao Yan; Peng Zhai; Lihua Zhang \\
  44 & \href{http://ieeexplore.ieee.org/document/9905530}{Convex Optimization for Trajectory Generation: A Tutorial On Generating Dynamically Feasible Trajectories Reliably And Efficiently} & Danylo Malyuta; Taylor P. Reynolds; Michael Szmuk; Thomas Lew; Riccardo Bonalli; Marco Pavone; Behçet Açıkmeşe \\
  45 & \href{https://journalofcloudcomputing.springeropen.com/articles/10.1186/s13677-023-00471-1}{A survey of Kubernetes scheduling algorithms} & Khaldoun Senjab; Sohail Abbas; Naveed Ahmed; Atta ur Rehman Khan \\
  46 & \href{https://www.mdpi.com/1424-8220/24/17/5551}{Auto-Scaling Techniques in Cloud Computing: Issues and Research Directions} & Saleha Alharthi; Afra Alshamsi; Anoud Alseiari; Abdulmalik Alwarafy \\
  47 & \href{https://ieeexplore.ieee.org/document/9837035/}{A Survey on Observability of Distributed Edge \& Container-Based Microservices} & Usman, M.; Ferlin, S.; Brunstrom, A.; Taheri, J. \\
  48 & \href{https://www.mdpi.com/2076-3417/15/17/9466}{Big Loop and Atomization: A Holistic Review on the Expansion Capabilities of Large Language Models} & Zefa Hu; Yi Huang; Junlan Feng; Chao Deng \\
  49 & \href{https://dl.acm.org/doi/10.1145/3736306}{HTTP Adaptive Streaming: A Review on Current Advances and Future Challenges} & Christian Timmerer; Hadi Amirpour; Farzad Tashtarian; Samira Afzal; Amr Rizk; Michael Zink; Hermann Hellwagner \\
  50 & \href{https://arxiv.org/abs/2209.05761}{A Survey on Mobile Edge Computing for Video Streaming: Opportunities and Challenges} & MUHAMMAD ASIF KHAN; EMNA BACCOUR; ZINA CHKIRBENE; AIMAN ERBAD; RIDHA HAMILA; MOUNIR HAMDI; MONCEF GABBOUJ \\
  51 & \href{https://www.mdpi.com/2079-9292/12/17/3563}{Evolution of Popularity and Multiaspectual Comparison of Widely Used Web Development Frameworks} & Jakub Swacha; Artur Kulpa \\
  52 & \href{https://www.mdpi.com/2227-7102/13/7/692}{Shaping the Future of Education: Exploring the Potential and Consequences of AI and ChatGPT in Educational Settings} & Simone Grassini \\
  53 & \href{https://www.mdpi.com/2227-7102/14/2/176}{The Evolving Classroom: How Learning Analytics Is Shaping the Future of Education and Feedback Mechanisms} & Hanan Sharif; Amara Atif \\
  54 & \href{https://www.mdpi.com/2078-2489/16/7/519}{AI-Integrated Scaffolding to Enhance Agency and Creativity in K-12 English Language Learners: A Systematic Review} & Molly Li; Joshua Wilson \\
  55 & \href{https://socialprotection-humanrights.org/resource/social-protection-in-the-cultural-and-creative-sectorcountry-practices-and-innovations/}{Social Protection in the Cultural and Creative Sector: Country Practices and Innovations} & Carlos Galian; Margherita Licata; Maya Stern-Plaza \\
  56 & \href{https://www.sciencedirect.com/science/article/pii/S2451958825000673}{Impacts of generative artificial intelligence on the future of labor market: A systematic review} & Nader Salari; Mahan Beiromvand; Amin Hosseinian-Far; Javad Habibi; Fateme Babajani; Masoud Mohammadi \\
  57 & \href{https://www.mdpi.com/2227-7099/12/2/35}{A Systematic Review of Industry 4.0 Technology on Workforce Employability and Skills: Driving Success Factors and Challenges in South Asia} & Md. Tota Miah; Szilvia Erdei-Gally; Anita Dancs; Mária Fekete-Farkas \\
  58 & \href{https://www.frontiersin.org/journals/education/articles/10.3389/feduc.2023.1334153/full}{A systematic review of the effectiveness of online learning in higher education during the COVID-19 pandemic period} & Wentao Meng; Lei Yu; Chen Liu; Nengchao Pan; Xiawen Pang; Yunyun Zhu \\
  59 & \href{https://www.mdpi.com/2227-7102/14/6/643}{ChatGPT in Teaching and Learning: A Systematic Review} & Duha Ali; Yasin Fatemi; Elahe Boskabadi; Mohsen Nikfar; Jude Ugwuoke; Haneen Ali \\
  60 & \href{https://www.mdpi.com/2227-7102/13/6/575}{A Scoping Review of School-Based Strategies for Addressing Anxiety, Intolerance of Uncertainty and Prediction in Autistic Pupils} & Anne Emerson; Debra Costley \\
  61 & \href{https://www.nature.com/articles/s41539-025-00320-7}{A systematic review of AI-driven intelligent tutoring systems (ITS) in K-12 education} & Angélique Létourneau; Marion Deslandes Martineau; Patrick Charland; John Alexander Karran; Jared Boasen; Pierre Majorique Léger \\
  62 & \href{https://pubmed.ncbi.nlm.nih.gov/36196475/}{Effectiveness of salt substitute on cardiovascular outcomes: A systematic review and meta-analysis} & Yi-Ching Tsai; Yen-Po Tsao; Chi-Jung Huang; Yen-Hsuan Tai; Yang-Chin Su; Chern-En Chiang; Shih-Hsien Sung; Chen-Huan Chen; Hao-Min Cheng \\
  63 & \href{https://www.mdpi.com/1422-0067/26/15/7400}{Mitochondrial Metabolism in T-Cell Exhaustion} & Fei Li; Yu Feng; Zesheng Yin; Yahong Wang \\
  64 & \href{https://pubmed.ncbi.nlm.nih.gov/37554723/}{The current state and future of T-cell exhaustion research} & Edward Jenkins; Toby Whitehead; Martin Fellermeyer; Simon J Davis; Sumana Sharma \\
  65 & \href{https://pubmed.ncbi.nlm.nih.gov/37108314/}{Zinc in Cardiovascular Functions and Diseases: Epidemiology and Molecular Mechanisms for Therapeutic Development} & Takafumi Hara; Emi Yoshigai; Takuto Ohashi; Toshiyuki Fukada \\
  66 & \href{https://www.mdpi.com/2076-2607/13/6/1410}{The Interplay Between the Gut Microbiota and Colorectal Cancer: A Review of the Literature} & Marco Cintoni; Marta Palombaro; Eleonora Zoli; Giuseppe D'Agostino; Gabriele Pulcini; Elena Leonardi; Pauline Raoul; Emanuele Rinninella; Flavio De Maio; Esmeralda Capristo; Antonio Gasbarrini; Maria Cristina Mele \\
  67 & \href{https://www.researchgate.net/publication/363416511_Impact_of_N6-methyladenosine_m6A_modification_on_immunity}{Impact of N6-methyladenosine (m⁶A) modification on immunity} & Raghda A. Elsabbagh; Mona Rady; Carsten Watzl; Khaled Abou-Aisha; Mohamed Z. Gad \\
  68 & \href{https://pubmed.ncbi.nlm.nih.gov/40773762/}{AI-Supported Shared Decision-Making (AI-SDM): Conceptual Framework} & Mohammed As'ad; Nawarh Faran; Hala Joharji \\
  69 & \href{https://www.mdpi.com/2227-9032/13/10/1205}{A Scoping Review of AI-Driven Digital Interventions in Mental Health Care: Mapping Applications Across Screening, Support, Monitoring, Prevention, and Clinical Education} & Yang Ni; Fanli Jia \\
  70 & \href{https://pubmed.ncbi.nlm.nih.gov/38519626/}{Digital twins for health: a scoping review} & Evangelia Katsoulakis; Qi Wang; Huanmei Wu; Leili Shahriyari; Richard Fletcher; Jinwei Liu; Luke Achenie; Hongfang Liu; Pamela Jackson; Ying Xiao; Tanveer Syeda-Mahmood; Richard Tuli; Jun Deng \\
  71 & \href{https://www.researchgate.net/publication/317401748_The_Psychology_of_Conspiracy_Theories}{The Psychology of Conspiracy Theories} & Karen M. Douglas; Robbie M. Sutton; Aleksandra Cichocka \\
  72 & \href{https://pubmed.ncbi.nlm.nih.gov/32438686/}{Clinical Features of Parkinson's Disease: The Evolution of Critical Symptoms} & Csaba Váradi \\
  73 & \href{https://doi.org/10.1093/bib/bbae461}{Knowledge Graphs for drug repurposing: a review of databases and methods} & Pablo Perdomo-Quinteiro; Alberto Belmonte-Hernández \\
  74 & \href{https://datascience.codata.org/articles/10.5334/dsj-2021-003}{A Review of Open Research Data Policies and Practices in China} & Lili Zhang; Robert R. Downs; Jianhui Li; Liangming Wen; Chengzan Li \\
  75 & \href{https://www.sciencedirect.com/org/science/article/pii/S2352133324000104}{Is a New Chinese Literary History Possible? A Critical Investigation of The Cambridge History of Chinese Literature} & Shen Yifan \\
  76 & \href{https://www.cambridge.org/core/journals/journal-of-global-history/article/special-issue-introduction-towards-a-global-history-of-international-organizations-and-decolonization/066AFEF8CEA16E7F776CC7CFF72927CC}{Special issue introduction: Towards a global history of international organizations and decolonization} & Eva-Maria Muschik \\
  77 & \href{https://hal.science/hal-05097583v1/document}{Transsexual Surgery in Egypt or the Suspicion of Homosexuality} & Corinne Fortier \\
  78 & \href{https://www.mdpi.com/2076-0752/7/4/56}{Anime in Academia: Representative Object, Media Form, and Japanese Studies} & Jaqueline Berndt \\
  79 & \href{https://jll.pitt.edu/ojs/JLL/article/view/129}{Toward More Inclusive Japanese Language Education: Incorporating an Awareness of Gender and Sexual Diversity among Students} & Jotaro Arimori \\
  80 & \href{https://www.mdpi.com/2076-0787/5/1/7}{Decolonization of Trauma and Memory Politics: Insights from Eastern Europe} & Dovile Budryte \\
  81 & \href{https://www.mdpi.com/2414-4088/3/2/39}{A Review of Augmented Reality Applications for History Education and Heritage Visualisation} & Jennifer Challenor; Minhua Ma \\
  82 & \href{https://journals.plos.org/plosone/article?id=10.1371/journal.pone.0311436}{Searching for the heritage of the Second Sino-Japanese War: A study on the site selection strategy of the defence industrial buildings of the National Resources Commission (1937–1945)} & Yangjie Wu \\
  83 & \href{https://www.frontiersin.org/journals/psychology/articles/10.3389/fpsyg.2023.1238272/full}{Collective memory: between individual systems of consciousness and social systems} & Jean-François Orianne; Francis Eustache \\
  84 & \href{https://www.mdpi.com/2227-7099/9/1/25}{The Policy Framework of Natural Resource Management in Oil-Dependence Countries} & Basem Ertimi; Tamat Sarmidi; Norlin Khalid; Mohd Helmi Ali \\
  85 & \href{https://www.mdpi.com/2076-0752/9/1/2}{'A Continuous Retrial': Trans/national Memory in Chinese and Japanese Tribunal Films} & Amanda Weiss \\
  86 & \href{https://pubs.aip.org/aip/jap/article/136/17/171101/3318627/Selecting-alternative-metals-for-advanced}{Selecting alternative metals for advanced interconnects} & Jean-Philippe Soulié; Kiroubanand Sankaran; Benoit Van Troeye; Alicja Leśniewska; Olalla Varela Pedreira; Herman Oprins; Gilles Delie; Claudia Fleischmann; Lizzie Boakes; Cédric Rolin; Lars-Åke Ragnarsson; Kristof Croes; Seongho Park; Johan Swerts; Geoffrey Pourtois; Zsolt Tőkei; Christoph Adelmann \\
  87 & \href{https://advanced.onlinelibrary.wiley.com/doi/10.1002/advs.202207321}{Materials Quest for Advanced Interconnect Metallization in Integrated Circuits} & Jun Hwan Moon; Eunjin Jeong; Seunghyun Kim; Taesoon Kim; Eunsoo Oh; Keun Lee; Hauk Han; Young Keun Kim \\
  88 & \href{https://www.mdpi.com/2079-4991/12/21/3845}{Recent Progress in Contact Engineering of Field-Effect Transistor Based on Two-Dimensional Materials} & Jialei Miao; Xiaowei Zhang; Ye Tian; Yuda Zhao \\
  89 & \href{https://www.mdpi.com/1911-8074/14/9/422}{The Determinants of PayTech's Success in the Mobile Payment Market—The Case of BLIK} & Joanna Błach; Monika Klimontowicz \\
  90 & \href{https://www.mdpi.com/2072-666X/13/8/1332}{SRAM Cell Design Challenges in Modern Deep Sub-Micron Technologies: An Overview} & Waqas Gul; Maitham Shams; Dhamin Al-Khalili \\
  91 & \href{https://www.mdpi.com/1424-8220/24/1/16}{Design of High-Speed, Low-Power Sensing Circuits for Nano-Scale Embedded Memory} & Sangheon Lee; Gwanwoo Park; Hanwool Jeong \\
  92 & \href{https://www.mdpi.com/2073-4441/13/9/1252}{Role of Models in the Decision-Making Process in Integrated Urban Water Management: A Review} & Leila Mosleh; Masoud Negahban-Azar \\
  93 & \href{https://www.mdpi.com/1424-8220/24/3/740}{An In-Depth Study of Vibration Sensors for Condition Monitoring} & Ietezaz Ul Hassan; Krishna Panduru; Joseph Walsh \\
  94 & \href{https://journals.plos.org/water/article?id=10.1371/journal.pwat.0000127}{Household, neighbourhood and service provider risk factors for piped drinking-water intermittency in urban and peri-urban Zambia: A cross-sectional analysis} & Mair L. H. Thomas-Possee; Andrew A. Channon; Robert E. S. Bain; James A. Wright \\
  95 & \href{https://jeas.springeropen.com/articles/10.1186/s44147-024-00546-z}{A review of emerging hydroforming technologies: design considerations, parametric studies, and recent innovations} & Satish Chinchanikar; Harsh Mulik; Param Varude; Sameer Atole; Neha Mundada \\
  96 & \href{https://www.mdpi.com/2076-3417/13/3/1687}{A Review of the High-Mix, Low-Volume Manufacturing Industry} & Zhi Lon Gan; Siti Nurmaya Musa; Hwa Jen Yap \\
  97 & \href{https://www.mdpi.com/2076-0752/11/5/90}{The Influencers: Van Gogh Immersive Experiences and the Attention-Experience Economy} & Kate Mondloch \\
  98 & \href{https://jjs.libraries.rutgers.edu/index.php/jjs/article/view/289}{We Have No More Creators: Mary Lou Williams Performs the Jazz Canon} & Sarah Caissie Provost \\
  99 & \href{https://www.cogitatiopress.com/mediaandcommunication/article/view/9356}{Televisuality on a Global Scale: Netflix's Local-Language Strategy} & Frédérique Khazoom \\
  100 & \href{https://arxiv.org/abs/2105.09266}{Copyright in Generative Deep Learning} & Giorgio Franceschelli; Mirco Musolesi \\
  101 & \href{https://www.mdpi.com/2078-2489/12/9/367}{Game Design as an Autonomous Research Subject} & Pedro Pinto Neves; Nelson Zagalo \\
  102 & \href{https://www.frontiersin.org/journals/sports-and-active-living/articles/10.3389/fspor.2021.640362/full}{The New Frontier of Esports and Gaming: A Scoping Meta-Review of Health Impacts and Research Agenda} & Sarah Kelly; Janni Leung \\
  103 & \href{https://www.mdpi.com/2032-6653/15/7/305}{Distribution of the Burden of Proof in Autonomous Driving Tort Cases: Implications of the German Legislation for China} & Zhihua Chen; Qianyi Cai; Hanbing Wei \\
  104 & \href{https://bmcmedethics.biomedcentral.com/articles/10.1186/s12910-015-0074-0}{Historical development and current status of organ procurement from death-row prisoners in China} & Kirk C Allison; Arthur Caplan; Michael E Shapiro; Charl Els; Norbert W Paul; Huige Li \\
  105 & \href{https://iopn.library.illinois.edu/journals/jams/article/view/808}{A Survey of the Story Elements of Isekai Manga} & Dr. Paul S. Price \\
  106 & \href{https://www.mdpi.com/2076-0752/14/4/85}{IP Adaptation Strategies in Film: A Case Study of Ne Zha 2 (2025)} & Aixin Chen; Haodong Gu \\
  107 & \href{https://link.springer.com/article/10.1007/s40864-024-00231-7}{Urban Rail Transit in China: Progress Report and Analysis (2015–2023)} & Kai Lu; Lei Zhang; Shen Li; Yunping Huang; Xiang Ding; Jingnan Hao; Siqi Huang; Xiaojuan Li; Fang Lu; Hongwei Zhang \\
  108 & \href{https://link.springer.com/article/10.1007/s40534-022-00281-2}{Pantograph-catenary electrical contact system of high-speed railways: recent progress, challenges, and outlooks} & Guangning Wu; Keliang Dong; Zhilei Xu; Song Xiao; Wenfu Wei; Huan Chen; Jie Li; Zhanglin Huang; Jingwei Li; Guoqiang Gao; Guozheng Kang; Chuanjun Tu; Xingyi Huang \\
  109 & \href{https://www.mdpi.com/1424-8220/17/6/1457}{Towards the Internet of Smart Trains: A Review on Industrial IoT-Connected Railways} & Paula Fraga-Lamas; Tiago M. Fernández-Caramés; Luis Castedo \\
  110 & \href{https://pubmed.ncbi.nlm.nih.gov/32471813/}{Physical activity and health in Chinese children and adolescents: expert consensus statement (2020)} & Peijie Chen; Dengfeng Wang; Hongbing Shen; Lijuan Yu; Qian Gao; Lijuan Mao; Fan Jiang; Yaojia Luo; Minhao Xie; Yong Zhang; Lianshi Feng; Feng Gao; Yuling Wang; Yu Liu; Chunyan Luo; George P Nassis; Peter Krustrup; Barbara E Ainsworth; Peter A Harmer; Fuzhong Li \\
  111 & \href{https://peerj.com/articles/17097/}{Physical activity and sedentary behavior among school-going adolescents in low- and middle-income countries: insights from the global school-based health survey} & Hui Li; Wenyu Zhang; Jin Yan \\
  112 & \href{https://www.mdpi.com/1424-8220/19/14/3160}{Video Activity Recognition: State-of-the-Art} & Itsaso Rodríguez-Moreno; José María Martínez-Otzeta; Basilio Sierra; Igor Rodriguez; Ekaitz Jauregi \\
  113 & \href{https://www.mdpi.com/2076-3417/15/3/1344}{AI-Driven Innovations in Software Engineering: A Review of Current Practices and Future Directions} & Mamdouh Alenezi; Mohammed Akour \\
  114 & \href{https://www.oecd.org/en/publications/competition-in-the-provision-of-cloud-computing-services_595859c5-en.html}{COMPETITION IN THE PROVISION OF CLOUD COMPUTING SERVICES} & OECD \\
  115 & \href{https://www.mdpi.com/2077-1444/16/3/383}{In Search of Qi Immortality: A Study of Heshanggong's Commentary on the Daodejing} & Jenny Hung \\
  116 & \href{https://www.mdpi.com/2077-1444/14/7/827}{Phenomenology of Quranic Corporeality and Affect: A Concrete Sense of Being Muslim in the World} & Valerie Gonzalez \\
  117 & \href{https://www.mdpi.com/2077-1444/16/2/184}{Is Emptiness Non-Empty? Jizang's Conception of Buddha-Nature} & Jenny Hung \\
  118 & \href{https://www.degruyterbrill.com/document/doi/10.1515/opth-2020-0012/html}{Mothers of a Nation: How Motherhood and Religion Intermingle in the Hebrew Bible} & Claudia D. Bergmann \\
  119 & \href{https://journals.plos.org/climate/article?id=10.1371/journal.pclm.0000266}{Effects and perceptions of weather, climate, and climate change on outdoor recreation and nature-based tourism in the United States: A systematic review} & Emily J. Wilkins; Lydia Horne \\
  120 & \href{https://journals.plos.org/plosone/article?id=10.1371/journal.pone.0293669}{Tourist distribution in Northern Mediterranean Basin countries: 2004–2020} & Sabri Öz; Adnan Veysel Ertemel; Pınar Başar; Cemil Can Çoktuğ \\
  121 & \href{https://www.mdpi.com/2071-1050/14/7/4295}{The Causality Analysis of Airports and Regional Economy: Empirical Evidence from Jiangsu Province in China} & Yang Bai; Cheng-Lung Wu \\
  122 & \href{https://www.mdpi.com/2071-1050/15/18/13544}{Resilient and Sustainable Housing Models against Climate Change: A Review} & Michelle A. Ruíz; Yazmin L. Mack-Vergara \\
  123 & \href{https://www.mdpi.com/2413-8851/9/4/101}{Trends, Methods, Drivers, and Impacts of Housing Informalities (HI): A Systematic Literature Review} & Rim Mrani; Hassan Radoine; Jérôme Chenal; Alanda Kamana \\
  124 &  &  \\
  125 & \href{https://journals.plos.org/plosone/article?id=10.1371/journal.pone.0294340}{Educational outcomes of recess in elementary school children: A mixed-methods systematic review} & Erin K. Howie; Kristi L. Perryman; Joseph Moretta; Laura Cameron \\
  126 & \href{https://pubmed.ncbi.nlm.nih.gov/37184735/}{24-h Movement Guidelines and Overweight and Obesity Indicators in Toddlers, Children and Adolescents: A Systematic Review and Meta-Analysis} & Adilson Marques; Rodrigo Ramirez-Campillo; Élvio Gouveia; Gerson Ferrari; Riki Tesler; Priscila Marconcin; Vânia Loureiro; João Martins; Hugo Sarmento \\
  127 & \href{https://www.frontiersin.org/journals/psychology/articles/10.3389/fpsyg.2024.1410462/full}{Social and ethical impact of emotional AI advancement: the rise of pseudo-intimacy relationships and challenges in human interactions} & Jie Wu \\
  128 & \href{https://pubmed.ncbi.nlm.nih.gov/38840697/}{Discussion on protein recommendations for supporting muscle and bone health in older adults: a mini review} & Inge Groenendijk; Lisette C. P. G. M. de Groot; Inge Tetens; Pol Grootswagers \\
  129 & \href{https://www.mdpi.com/2306-5710/10/3/49}{Low-Alcohol and Nonalcoholic Wines: From Production to Cardiovascular Health, along with Their Economic Effects} & Paula Silva \\
  130 & \href{https://pubmed.ncbi.nlm.nih.gov/30884834/}{Non-Nutritive Sweeteners and Their Implications on the Development of Metabolic Syndrome} & Iryna Liauchonak; Bessi Qorri; Fady Dawoud; Yatin Riat; Myron R. Szewczuk \\
  131 & \href{https://pubmed.ncbi.nlm.nih.gov/38132429/}{Profile of Orthodontic Use across Demographics} & Man Hung; Golnoush Zakeri; Sharon Su; Amir Mohajeri \\
  132 & \href{https://pubmed.ncbi.nlm.nih.gov/38592701/}{Management of Melasma: Laser and Other Therapies—Review Study} & Badea Jiryis; Ohad Toledano; Emily Avitan-Hersh; Ziad Khamaysi \\
\end{longtable}

\section{Source Article Leakage Rate}
\label{sec:source-leakage}
As mentioned in the main text, we attempted to prevent the model from accessing source articles solely through the prompt, but we could not fully eliminate the possibility of the model viewing the source articles. Therefore, during the evaluation, we conducted a secondary check of the generated reports and calculated the leakage rate of the articles. The results are shown in the table below:

\begin{table}[ht]
\centering
\small
\begin{tabular}{lccccccccc}
\toprule
\textbf{Model} & \textbf{Qwen} & \textbf{Gemini-2.5} & \textbf{OpenAI} & \textbf{Tongyi} & \textbf{Gemini-3} & \textbf{Doubao} & \textbf{Perplexity} & \textbf{Grok} & \textbf{Avg.} \\
\midrule
Leakage Rate (\%) & 13.20 & 4.54 & 2.67 & 2.32 & 1.90 & 1.64 & 0.85 & 0.24 & 3.64 \\
\bottomrule
\end{tabular}
\caption{Source Article Leakage Rate by Model}
\label{tab:leakage_rate}
\end{table}

The leakage rate for each model is non-zero. However, with the exception of Qwen, the leakage rates for the other models are all within 5\%. Overall, the leakage rate is relatively low, indicating that the blocked list in the prompt has been somewhat effective in limiting access to source articles. The specific list of source articles is provided in Appendix~\ref{sec:source-article-list}.

\section{Prompt}
We release the prompts used for task/rubric generation and for evaluation (rubric scoring) to enable reproducibility and independent analysis.

\subsection{Tasks and Rubrics Generation Prompt}
\begin{tcolorbox}[
  breakable,
  colback=porcelain,
  colframe=gray!75!black,
  left=6pt, right=6pt, top=6pt, bottom=6pt,
  fontupper=\small
]

I want to construct prompts that test current large language models' deep research capability (``deepresearch''). We consider deepresearch to consist of three components: information recall, analysis, and presentation. Therefore, we will score along these three dimensions.

Concretely, we will take a portion of content from an existing article written by a human expert. You can regard this extracted portion as the ``reference answer.'' Next, we reverse-engineer a question from the ``reference answer,'' i.e., we summarize a task. The core of this task is the recall of information and data from the open internet. The model's goal should be to, given this task, retrieve information across the entire internet and produce content that is as similar as possible to the human article---or to a specific part of it (e.g., a subsection or a table). In addition, you must generate a rubric from this ``reference answer,'' which will be used to evaluate the model's output. The rubric has three dimensions: \texttt{info\_recall}, \texttt{analysis}, and \texttt{presentation}. 

\begin{itemize}
\item \texttt{info\_recall} evaluates whether the model has retrieved the factual information, data, and literature relevant to the task, and whether the retrieved information is correct.
\item \texttt{analysis} evaluates whether the model has appropriately analyzed the retrieved information to reach the conclusions required by the task, and whether those conclusions are similar to those in the human article (the difference between \texttt{info\_recall} and \texttt{analysis} is analogous to the difference between facts and viewpoints).
\item \texttt{presentation} evaluates whether the model's output format is correct, whether it meets the user's formatting requirements, and whether the content presentation is appropriate.
\end{itemize}

In your output, you must also include a field named \texttt{"blocked"}, whose content is the title and author(s) of the human reference article. This article must not be visible to the model when it generates answers.

Your final output must follow JSON format. See the example below:

\textbf{Example 1:}

\begin{lstlisting}[basicstyle=\ttfamily\footnotesize, breaklines=true, frame=none, numbers=none]
{
  "task": "Please complete the following around the museologist Peter van Mensch:\\n1. Outline his life trajectory, including birth information, education, academic appointments, and major roles in academic organizations.\\n2. Summarize his theoretical contributions to museology, covering the seven transitions he proposed, his classification of museum objects, the concepts of musealia and musealisation, and the key points of his museum management model.\\n3. Analyze his role in the international museology community, especially his academic connections with Zbynek Stransky and Ivo Maroevic, and his role in organizations such as ICOFOM.\\n4. Evaluate his impact on the disciplinization, professionalization, and socialization of museology, integrating his educational philosophy and practice.\\n5. Provide a complete list of his major works in chronological order, ensuring that no representative work is omitted.\\nThe output must be structured into four parts: Biography, Points of view on museology, Main Works, and Analysis.",
  "rubric": {
    "info_recall": [
      "Explicitly state that Peter van Mensch was born in 1947 and is Dutch.",
      "Explicitly state that Peter van Mensch began teaching at the Reinwardt Academy in Amsterdam in 1978.",
      "Explicitly state that Peter van Mensch became a senior lecturer in theoretical museology in 1982.",
      "Explicitly state that Peter van Mensch became a professor of cultural heritage in 2006.",
      "Explicitly state that Peter van Mensch retired in 2011.",
      "Explicitly state that Peter van Mensch completed his PhD at the University of Zagreb.",
      "Explicitly state that Peter van Mensch’s PhD supervisor/collaborator included Ivo Maroević.",
      "Explicitly state that Peter van Mensch’s doctoral research was influenced by the theory of Zbyněk Z. Stránský.",
      "Explicitly state that Peter van Mensch served as ICOFOM president from 1989 to 1993.",
      "Explicitly state that Peter van Mensch long served as a board member or advisor of ICOFOM.",
      "Explicitly state that the title of Peter van Mensch’s dissertation is “Towards a Methodology of Museology,” completed in 1992.",
      "Explicitly state that Peter van Mensch established a teaching system combining theory and practice at the Reinwardt Academy.",
      "Explicitly list the seven transitions proposed by Peter van Mensch: from object-centric to interpretation-centric.",
      "Explicitly list the seven transitions: expansion of the concept of museum objects to include intangible heritage and technical objects.",
      "Explicitly list the seven transitions: emphasis on contextual preservation, valuing original environments.",
      "Explicitly list the seven transitions: decentralization, moving from national museums toward local and community museums.",
      "Explicitly list the seven transitions: from “museum of things” to “museum of ideas,” with exhibitions emphasizing concept communication.",
      "Explicitly list the seven transitions: managerial rationalization, employing systematic management models.",
      "Explicitly list the seven transitions: expansion of the scope of musealisation, including cultural and commercial institutions within its ambit.",
      "Explicitly list Peter van Mensch’s classification of museum objects: artifacts.",
      "Explicitly list the classification: documents.",
      "Explicitly list the classification: books.",
      "Explicitly list the classification: buildings.",
      "Explicitly list the classification: living entities in museum environments.",
      "Explicitly state that Peter van Mensch used the term musealia to refer to museum objects.",
      "Explicitly state that Peter van Mensch used the term musealisation to refer to the process of becoming musealized.",
      "Explicitly list his museum management model: an input–throughput–output process perspective.",
      "Explicitly list the management model’s “internal coupling” (coordination among administration, research, collections, and dissemination).",
      "Explicitly list the management model’s “external coupling” (interaction with audiences, communities, and networks).",
      "Explicitly list the work: 1983 “Natural history museums – New directions,” Reinwardt Studies in Museology 1, pp. 55–63.",
      "Explicitly list the work: 1984 “Society – object – museology,” ICOFOM Study Series, 6, 18–23.",
      "Explicitly list the work: 1984 “Collecting today for tomorrow,” ICOFOM Study Series, 7, 29–32.",
      "Explicitly list the work: 1985 “Museological relevance of management techniques,” in Management needs of museum personnel, Reinwardt Studies in Museology 5, pp. 9–15.",
      "Explicitly list the work: 1985 “Museums and authenticities: provocative thoughts,” ICOFOM Study Series, 8, 13–20.",
      "Explicitly list the work: 1985 “Towards a typology of copies,” ICOFOM Study Series, 8, 123–126.",
      "Explicitly list the work: 1985 “Originals and Substitutes in Museums,” ICOFOM Study Series, 9, 45–50.",
      "Explicitly list the work: 1986 “Museology and identity,” ICOFOM Study Series, 10, 201–209.",
      "Explicitly list the work: 1986 “Museology and identity: comments and views,” ICOFOM Study Series, 11, 37–39.",
      "Explicitly list the work: 1987 “Museologists in a train to Helsinki,” ICOFOM Study Series, 13, 47–51.",
      "Explicitly list the work: 1987 “Museums in movement,” ICOFOM Study Series, 12, 17–20.",
      "Explicitly list the work: 1987 “Musées en mouvement,” ICOFOM Study Series, 12, 25–29.",
      "Explicitly list the work: 1988 “Museology and museums,” ICOM News, 41(3), 5–10.",
      "Explicitly list the work: 1988 “What contribution has museology to offer to the developing countries?”, ICOFOM Study Series, 14, 181–185.",
      "Explicitly list the work: 1989 “Professionalizing the Muses. The museum profession in motion,” Amsterdam: AHA Books.",
      "Explicitly list the work: 1989 “Forecasting – a museological tool?”, ICOFOM Study Series, 16, 175–178.",
      "Explicitly list the work: 1990 “Methodological museology; or, towards a theory of museum practice,” in Pearce, S. (Ed.), New Research in Museum Studies 1, pp. 141–157.",
      "Explicitly list the work: 1990 “Museology and the environment,” ICOFOM Study Series, 17, 13–14.",
      "Explicitly list the work: 1991 “ICOFOM ’91 symposium: the language of exhibitions,” ICOFOM Study Series, 19, 11–13.",
      "Explicitly list the work: 1992 “Museological research,” Museological News, 15, 8.",
      "Explicitly list the work: 1992 “Towards a Methodology of Museology,” PhD dissertation, University of Zagreb.",
      "Explicitly list the work: 1992 “Museological research. Current affairs in museology,” ICOFOM Study Series, 21, 3–4.",
      "Explicitly list the work: 1992 “Museological research,” ICOFOM Study Series, 21, 19–33.",
      "Explicitly list the work: 1992 “Summaries of ICOFOM symposia 1976–1991,” ICOFOM Study Series, 21, 97–101.",
      "Explicitly list the work: 1993 “Towards museums for a new century,” ICOFOM Study Series, 22, 15–19.",
      "Explicitly list the work: 1993 “ICOFOM and the basic parameters in museology,” ICOFOM Study Series, 22, 101–103.",
      "Explicitly list the work: 1993 “Master the art of museum studies in Amsterdam,” ICOFOM Study Series, 22, 117–118.",
      "Explicitly list the work: 1994 “Museum analysis model – an outline,” in Theoretical Museology, pp. 183–184, Amsterdam: Reinwardt Academy.",
      "Explicitly list the work: 1994 “The characteristics of exhibitions,” in Theoretical Museology, pp. 185–192, Amsterdam: Reinwardt Academy.",
      "Explicitly list the work: 1994 “Towards a methodology of museology,” ICOFOM Study Series, 23, 59–69.",
      "Explicitly list the work: 1994 “Object – document? Summary and final remarks,” ICOFOM Study Series, 23, 195–203.",
      "Explicitly list the work: 1995 “Magpies on Mount Helicon?”, ICOFOM Study Series, 25, 133–138.",
      "Explicitly list the work: 2000 “Nieuwe Visies voor de 21ste eeuw,” Museumvisie, 24(1), ii–ix.",
      "Explicitly list the work: 2000 “Museology as a profession,” ICOM Study Series, 8, 20–21.",
      "Explicitly list the work: 2001 “Museum Studies in the Netherlands,” in Scaltsa, M. (Ed.), Museology Towards the 21st Century, pp. 146–149.",
      "Explicitly list the work: 2003 “The characteristics of exhibitions,” Museum Aktuell, 92, 3980–3985.",
      "Explicitly list the work: 2003 “Convergence and divergence. Museums of science and technology in historical perspective,” in Simard, C. (Ed.), pp. 342–352.",
      "Explicitly list the work: 2004 “Museology and management: Enemies or friends? Current tendencies” (ends here as in the original)."
    ],
    "analysis": [
      "Explicitly explain that he chose the University of Zagreb for PhD study under the influence of Stránský and Ivo Maroević’s school, aiming to systematize Central European theoretical museology.",
      "Explicitly explain that he designed courses combining theory and practice at the Reinwardt Academy to enhance the systematization and operability of professional training in museology.",
      "Explicitly explain that his introduction of “new museology” responded to the global trend in the 1970–80s toward social function and audience orientation.",
      "Explicitly explain that his introduction of “social museology” emphasized social inclusion, participatory learning, and community collaboration.",
      "Explicitly explain that his proposal of “metamuseology” aimed to promote reflexive positioning of museology and to clarify disciplinary boundaries.",
      "Explicitly explain that during his 1989–1993 ICOFOM presidency he shifted the organization’s focus from experience sharing to theoretical research and terminology standardization.",
      "Explicitly explain that the shift from object-centric to interpretation-centric exhibitions was to strengthen the educational function of museums.",
      "Explicitly explain the relation between expanding object concepts and the entrance of intangible heritage into museological research.",
      "Explicitly explain the relation between decentralization and the rise of community/ecomuseums.",
      "Explicitly explain the relation between the shift to “museums of ideas” and the emergence of concept communication and theme curation.",
      "Explicitly explain that introducing the terms musealia and musealisation aimed to unify disciplinary language and avoid conflating objects with processes.",
      "Explicitly explain that combining musealisation with semiotics means objects gain new meanings through display once inside the museum.",
      "Explicitly explain that “internal coupling” in his management model aims to enhance coordination across administration, research, collections, and dissemination.",
      "Explicitly explain that “external coupling” aims to foster interaction between museums and communities/publics.",
      "Explicitly explain that developing management theory within ICTOP aimed to cultivate museum professionals with systems thinking.",
      "Explicitly explain why he is categorized as a “second-generation museologist”: inheriting Stránský’s scientific framework and extending the methodology.",
      "Explicitly explain that his educational contributions at the Reinwardt Academy directly advanced the professional training of museum personnel in Europe.",
      "Explicitly explain that emphasizing public participation in his theoretical model aimed to promote the socialization of museums.",
      "Explicitly explain the mechanism by which he leveraged ICOFOM to expand Central European theory into a global, cross-regional dialogue network."
    ],
    "presentation": [
      "The answer must be divided into four parts: Biography, Points of view on museology, Main Works, and Synthesis/Analysis.",
      "In the Main Works section, the works must be strictly presented in chronological order and cover all items listed in the rubric.",
      "Each section must have clear subheadings; language must be precise and logic coherent."
    ]
  },
  "blocked": [
      "A History of Museology, Bruno Brulon Soares"
  ]
}
\end{lstlisting}

\textbf{Example 2:}

\begin{lstlisting}[basicstyle=\ttfamily\footnotesize, breaklines=true, frame=none, numbers=none]
{
  "task": "Conduct a study on SRAM Sensing Circuits for Offset Reduction, and produce two parts:\\nA) Technical survey: in the order of appearance, enumerate each type of sense amplifier (SA) by name, core offset-reduction mechanism (why it reduces offset), key structural points (including component makeup / whether extra generators are required), required control signals, and major costs/risks (e.g., speed degradation / power & area increase / mismatch sensitivity / bit-line floating). Also add a 1-2 sentence trade-off summary for each design.\\nB) Comparison table: produce a table titled Comparison of SRAM sensing circuit designs with fixed columns [Structure | Offset Reduction Technique | Components | Control Signals | Limitations]; rows must fully cover: VLSA, CLSA, STSA, VBSTSA, VTSA, HYSA-QZ, TMSA, VTS-SA, CSAOC, BP2SP, CCN-PP, OCCSA, SAOC. Both parts must avoid vague wording-information must be populated item by item.\\nOutput format: first the Technical Survey (list in the specified order), then the Comparison Table (with the required column headers and all rows).",
  "rubric": {
    "info_recall": [
      "In the technical survey, explicitly provide the full name of STSA (Schmitt trigger–based SA).",
      "In the survey, explicitly state STSA’s core offset-reduction mechanism: “Driving internal nodes of VLSA (ZT and ZC).”",
      "In the survey, explicitly state STSA’s main cost includes “Speed degradation due to stack.”",
      "Explicitly provide the full name of VBSTSA (Voltage-Boosted STSA).",
      "Explicitly state VBSTSA’s core mechanism: “STSA + Negative Boosting VSS.”",
      "Explicitly state VBSTSA’s main cost includes “Requires an NVG (negative voltage generator), incurring power/area overhead.”",
      "Explicitly provide the name of VTSA (Variation-Tolerant SA).",
      "Explicitly state VTSA’s core mechanism: “Pre-charging SOT and SOC to SLT and SLC in CLSA (pre-charging internal/output nodes).”",
      "Explicitly state VTSA’s main cost includes “Speed degradation due to stack.”",
      "Explicitly provide the name of HYSA-QZ (Hybrid SA – QZ).",
      "Explicitly state HYSA-QZ’s core mechanism: “Pre-charging output nodes and internal nodes of CLSA.”",
      "Explicitly state HYSA-QZ’s main cost includes “Speed degradation due to stack.”",
      "Explicitly provide the name of TMSA (Threshold-Matching SA).",
      "Explicitly state TMSA’s core mechanism: “Capturing Vth of pull-down nFETs through paired capacitors.”",
      "Explicitly state TMSA’s main cost includes “Capacitor mismatch and cap power/area overhead.”",
      "Explicitly provide the name of VTS-SA (Variation-Tolerant Small-Signal SA / Voltage-Transfer type).",
      "Explicitly state VTS-SA’s core mechanism: “Capturing trip points of cross-coupled inverters with input acceptation via a coupling capacitor pair.”",
      "Explicitly state VTS-SA’s main cost includes “Capacitor mismatch and power/area overhead.”",
      "Explicitly provide the name of CSAOC (Capacitor Shared Offset Cancellation).",
      "Explicitly state CSAOC’s core mechanism: “Capturing trip points of cross-coupled inverters via a single capacitor.”",
      "Explicitly state CSAOC’s main cost includes “Many switches; complex control-signal circuitry.”",
      "Explicitly provide the name of BP2SP (Bit-line Pre-charge & Pre-amplifying Switching pFET).",
      "Explicitly state BP2SP’s core mechanism: “Capturing Vth of pre-amplifying pFET pair at BL pre-charge.”",
      "Explicitly state BP2SP’s main cost includes “Bit-line floating, unstable pre-charge level, power/area overhead.”",
      "Explicitly provide the name of CCN-PP (Cross-Coupled nFET Pre-amplifier & Pre-charge).",
      "Explicitly state CCN-PP’s core mechanism: “Pre-amplifying BL via cross-coupled nFET pair while capturing Vth with boosted VDD.”",
      "Explicitly state CCN-PP’s main cost includes “Bit-line floating, power/area overhead.”",
      "Explicitly provide the name of OCCSA (Offset-Cancelled Current-Mode SA / Capturing Vth of MUX nFETs at BL pre-charge).",
      "Explicitly state OCCSA’s core mechanism: “Capturing Vth of MUX nFETs at BL pre-charge.”",
      "Explicitly state OCCSA’s main cost includes “Requires additional Vprebl voltage source; complexity due to unsynchronized MUX signals.”",
      "Explicitly provide the name of SAOC (Sense-Amplifier Offset Cancellation).",
      "Explicitly state SAOC’s core mechanism: “Capturing Vth of input pFETs at SA pre-charge.”",
      "Explicitly state SAOC’s main cost includes “N1/N2 mismatch risk; added complexity in control-signal circuitry.”",
      "The comparison table must include the columns: Structure, Offset Reduction Technique, Components, Control Signals, Limitations.",
      "Provide a complete row for VLSA: Offset Reduction Technique “–”, Components “7 TR”, Control Signals “PCB, SAE”, Limitations “Large Vos”.",
      "Provide a complete row for CLSA: Offset Reduction Technique “–”, Components “9 TR”, Control Signals “PCB, SAE”, Limitations “Increased Vos due to additional TR pair”.",
      "Provide a complete row for STSA: Offset Reduction Technique “Driving internal nodes of VLSA (ZT and ZC)”, Components “11 TR”, Control Signals “PCB, SAE”, Limitations “Speed degradation due to stack”.",
      "Provide a complete row for VBSTSA: Offset Reduction Technique “STSA + Negative Boosting VSS”, Components “14 TR + NVG (share)”, Control Signals “PCB, SAE, BSTEN”, Limitations “Necessitating NVG (power/area cost)”.",
      "Provide a complete row for VTSA: Offset Reduction Technique “Pre-charging SOT and SOC to SLT and SLC in CLSA”, Components “9 TR”, Control Signals “PCB, SAE”, Limitations “Speed degradation due to stack”.",
      "Provide a complete row for HYSA-QZ: Offset Reduction Technique “Pre-charging output nodes and internal nodes of CLSA”, Components “11 TR”, Control Signals “PCB, SAE”, Limitations “Speed degradation due to stack”.",
      "Provide a complete row for TMSA: Offset Reduction Technique “Capturing Vth of pull-down nFETs through paired caps”, Components “11 TR + INV + Buffer + 2 C”, Control Signals “PCB, SAE”, Limitations “Capacitor mismatch; cap power/area overhead”.",
      "Provide a complete row for VTS-SA: Offset Reduction Technique “Capturing trip points of cross-coupled INVs via coupling cap pair”, Components “12 TR + 2 C”, Control Signals “EN, PCB, PRE, SAE”, Limitations “Capacitor mismatch; power/area overhead”.",
      "Provide a complete row for CSAOC: Offset Reduction Technique “Capturing trip points of cross-coupled inverters via single capacitor”, Components “16 TR + 1 C + 2 OR (shared)”, Control Signals “PCB, SAE, Φtrs, Φtrb”, Limitations “Many switches; complex control circuit”.",
      "Provide a complete row for BP2SP: Offset Reduction Technique “Capturing Vth of pre-amplifying pFET pair at BL pre-charge”, Components “6 TR + SA”, Control Signals “PCB, SAE”, Limitations “Bit-line floating; unstable pre-charge; power/area overhead”.",
      "Provide a complete row for CCN-PP: Offset Reduction Technique “Pre-amplifying BL via cross-coupled nFET pair, while capturing Vth with boosted VDD”, Components “4 TR + 2 C + Buffer + 1 TR + SA”, Control Signals “PCB, SAE, PBE”, Limitations “Bit-line floating; power/area overhead”.",
      "Provide a complete row for OCCSA: Offset Reduction Technique “Capturing Vth of MUX nFETs at BL pre-charge”, Components “7 TR”, Control Signals “PCB, SAE”, Limitations “Additional Vprebl generator; unsynchronized MUX signal”.",
      "Provide a complete row for SAOC: Offset Reduction Technique “Capturing Vth of input pFETs at SA pre-charge”, Components “11 TR”, Control Signals “PCB, SAE, OCEN”, Limitations “N1/N2 mismatch; control-signal circuit complexity”."
    ],
    "analysis": [
      "Provide mechanism-to-cost causal explanations for STSA: how driving internal nodes effectively lowers the trigger threshold to suppress offset, and why stacked paths degrade speed.",
      "Provide mechanism-to-cost causal explanations for VBSTSA: why negative boosting VSS further reduces effective offset on top of STSA, and the area/power costs of the NVG.",
      "Provide mechanism-to-cost causal explanations for VTSA/HYSA-QZ: how pre-charging internal/output nodes changes initial bias to reduce comparison offset, and why stacking causes speed loss.",
      "Provide mechanism-to-cost causal explanations for TMSA: how paired-capacitor sampling of pull-down nFET Vth cancels mismatch and reduces effective VOS, and how capacitor mismatch and area/power overhead impact.",
      "Provide mechanism-to-cost causal explanations for VTS-SA/CSAOC: how capturing trip points of cross-coupled inverters and using coupling/shared capacitors amplifies small input differentials to lower input thresholds; also explain sensitivity to control phasing complexity and capacitor mismatch.",
      "Provide mechanism-to-cost causal explanations for BP2SP/CCN-PP: how “pre-charge as pre-amplification” lowers required ΔV_BL to improve initial discriminability; and point out risks of BL floating / unstable pre-charge / higher power.",
      "Conclude with selection guidance by application scenario: e.g., ultra-low Vmin/high robustness → prefer capacitor-compensation family (TMSA/VTS-SA/CSAOC) while accepting area/complexity; extreme speed → consider pre-amplifier family (BP2SP/CCN-PP) with strict timing/power management; low-power IoT/area-constrained → consider VTSA/HYSA-QZ with simpler control while accepting some offset/speed trade-offs."
    ],
    "presentation": [
      "Output the “Technical Survey” first and then the “Comparison Table,” clearly separated with subheadings.",
      "The survey must cover, in order: STSA, VBSTSA, VTSA, HYSA-QZ, TMSA, VTS-SA, CSAOC, BP2SP, CCN-PP, OCCSA, SAOC; each entry must include: name, mechanism, structural points, control signals, costs, and a one-sentence trade-off.",
      "The table must have columns Structure, Offset Reduction Technique, Components, Control Signals, Limitations, and must contain exactly 13 rows: VLSA, CLSA, STSA, VBSTSA, VTSA, HYSA-QZ, TMSA, VTS-SA, CSAOC, BP2SP, CCN-PP, OCCSA, SAOC. Any missing row/field is unacceptable.",
      "No vague phrases such as 'etc.' or 'non-exhaustive.' All required items must be explicitly stated."
    ]
  },
  "blocked": [
    "Design of High-Speed, Low-Power Sensing Circuits for Nano-Scale Embedded Memory by Sangheon Lee 1ORCID, Gwanwoo Park 1 and Hanwool Jeong"
  ]
}
\end{lstlisting}

\textbf{Requirements:}

\textit{Notes:}

\begin{enumerate}
\item The human passage will NOT be visible to the model being evaluated; the model will only see the task and must gather information online by itself.
\item The task you extract does not have to cover the entire article; you may select a portion (e.g., one subsection or one table)---especially the part most related to deepresearch (data recall and organization)---to form the task. The task may avoid overly subjective parts of the source article, since a model is unlikely to reproduce the exact same analysis.
\item For critical data (e.g., table data), the task must specify exactly how to organize it (e.g., what columns to include in a table). For ambiguous concepts, provide strict definitions and clarify category distinctions.
\item The task you generate should include both information gathering and qualitative analysis (or simple quantitative analysis). It must not be a purely discursive piece without incorporating additional sources, nor merely data collection without analysis, nor an overly complex analysis (e.g., requiring massive datasets or machine-learning algorithms).
\item When generating the task, mimic a typical user's tone: assume the user is not an expert and cannot name specific in-field methods/details; use natural language, not Markdown, to describe the task.
\item Pay attention to timeliness: for topics that change quickly (e.g., research or industry), the task must explicitly state that only information up to a specific date should be collected, aligned with the human article's timeframe.
\item Your rubrics must be as comprehensive as possible and fully cover the task; do not trim rubrics due to length.
\item For numeric data, you must verify each value via rubrics (if values are derived by calculation, you may specify an allowable error).
\item If rubrics become too many, shrink the scope of the task rather than cutting rubrics. Ultimately, rubrics should number between 50 and 200. Note: shrinking scope does not imply directly listing which entities/methods/companies to search; discovering and enumerating them is part of deep research capability.
\item Rubric items should be as detailed as possible; a rubric is itself an answer to what must be verified. Avoid broad items (e.g., ``Gives reasons for talent outflow from small cities''). Instead: ``States that a key reason for talent outflow from small cities is a mismatch in occupational structure.''
\item Rubric granularity must be as fine as possible and not further divisible. For example, do not write ``Covers Axitinib (a VEGFR inhibitor that alleviates hypoxia\ldots) and NADH removal (\ldots)'', but split into ``Mentions Axitinib,'' ``Explains that Axitinib's VEGFR-inhibition relieves hypoxia and improves immunotherapy efficacy,'' etc.
\item If a rubric references other studies/views, you must name the specific research/article title rather than author shorthand. For example: ``Explicitly cites the study `Continuous Influx of Genetic Material from Host to Virus Populations' published in PLoS Genetics in 2016 by Gilbert et al., stating that 4.8\% of viruses contain host sequences,'' rather than ``Gilbert (2016) reported\ldots''.
\item Rubric items should be mutually independent, i.e., each rubric can be used to evaluate output on its own.
\end{enumerate}

Below, you will be given a portion of a human-written article. Following the examples above, extract an appropriate task and rubrics for that article.

\texttt{<human\_passage>}

\texttt{\{paper\}}

\texttt{</human\_passage>}

Now, please begin your generation.

\end{tcolorbox}

\subsection{Rubric Scoring Prompt}

\begin{tcolorbox}[breakable, colback=gray!5!white, colframe=gray!75!black, left=6pt, right=6pt, top=6pt, bottom=6pt, fontupper=\small]

You will receive an article, a task, and a list of grading rubric items. Your job is to assess whether the article satisfies each rubric item, and provide a THREE-WAY score for EACH rubric item.

\textbf{Scoring rule per rubric item (strict):}

\begin{itemize}
\item Score = 1: The article clearly satisfies the rubric item AND the specific supporting sentence(s) do NOT cite any reference listed in ``blocked'' (match by title/urls). For numerical data, exact values must be explicitly listed and match the rubric.
\item Score = 0: The article does NOT mention this rubric item at all.
\item Score = -1: The article mentions this rubric item, BUT the supporting sentence(s) cite a blocked reference.
\end{itemize}

\textbf{For EACH rubric item, you MUST provide:}

\begin{enumerate}
\item ``score'': 1, 0, or -1
\item ``reason'': A brief explanation
\item ``evidence'': The specific supporting sentence(s) from the article (empty string if score is 0)
\end{enumerate}

The input format is:

\texttt{<input\_format>}

\begin{lstlisting}[basicstyle=\ttfamily\footnotesize, breaklines=true, frame=none, numbers=none]
{
    "task": "...",
    "rubric_items": ["rubric item 1", "rubric item 2", ...],
    "blocked": {
        "title": "...",
        "authors": ["...", "..."],
        "urls": ["...", "..."]
    }
}
\end{lstlisting}

\texttt{</input\_format>}

Your output MUST strictly follow this JSON format (no extra keys, and the rubric item text MUST match the input EXACTLY):

\texttt{<output\_format>}

\begin{lstlisting}[basicstyle=\ttfamily\footnotesize, breaklines=true, frame=none, numbers=none]
{
    "results": [
        {
            "rubric_item": "rubric item 1",
            "score": 1 or 0 or -1,
            "reason": "brief explanation",
            "evidence": "supporting sentence(s) from the article"
        },
        {
            "rubric_item": "rubric item 2",
            "score": 1 or 0 or -1,
            "reason": "brief explanation",
            "evidence": "supporting sentence(s) from the article"
        },
        ...
    ]
}
\end{lstlisting}

\texttt{</output\_format>}

CRITICAL: You MUST return results for ALL rubric items in the input, and the ``rubric\_item'' text MUST match the input text EXACTLY (character-level match).

\texttt{<passage>}

\texttt{\{paper\}}

\texttt{</passage>}

\texttt{<task\_and\_rubric>}

\texttt{\{rubric\}}

\texttt{</task\_and\_rubric>}

Now, please begin your generation

\end{tcolorbox}

\end{document}